\documentclass[lettersize,journal]{IEEEtran}

\usepackage{amsmath,amsfonts,amssymb}
\usepackage{algorithmic}
\usepackage{algorithm}
\usepackage{array}
\usepackage[caption=false,font=normalsize,labelfont=sf,textfont=sf]{subfig}
\usepackage{textcomp}
\usepackage{stfloats}
\usepackage{url}
\usepackage{verbatim}
\usepackage{graphicx}
\usepackage{cite}
\usepackage{multirow}
\usepackage{stfloats}

\usepackage{xcolor}
\begin{document}

\title{Embedded Heterogeneous Attention Transformer for Cross-lingual Image Captioning}

\author{
Zijie~Song,~\IEEEmembership{Student Member,~IEEE},
Zhenzhen~Hu$^\dagger$,~\IEEEmembership{Member,~IEEE}, 
Yuanen~Zhou,~\IEEEmembership{Student Member,~IEEE},
Ye~Zhao,~\IEEEmembership{Member,~IEEE}, 
Richang~Hong,~\IEEEmembership{Member,~IEEE},
Meng~Wang,~\IEEEmembership{Fellow,~IEEE}

\thanks{Zhenzhen Hu$^\dagger$ is the corresponding author.

Zijie Song, Zhenzhen Hu, Ye Zhao, Richang Hong and Meng Wang are with the School of Computer Science and Information Engineering, Hefei University of Technology, Hefei, 230601, China, (e-mail: zjsonghfut@gmail.com; huzhen.ice@gmail.com; zhaoye@hfut.edu.cn; hongrc.hfut@gmail.com; eric.mengwang@gmail.com).

Yuanen Zhou is with the Institute of Artificial Intelligence, Hefei Comprehensive National Science Center, Hefei, 230088, China (e-mail: y.e.zhou.hb@gmail.com.)}

}

\markboth{This article has been accepted for publication in IEEE Transactions on Multimedia (Early Access). \\ 
DOI 10.1109/TMM.2024.3384678}{}




\maketitle

\begin{abstract}

Cross-lingual image captioning is a challenging task that requires addressing both cross-lingual and cross-modal obstacles in multimedia analysis. The crucial issue in this task is to model the global and the local matching between the image and different languages. Existing cross-modal embedding methods based on the transformer architecture oversee the local matching between the image region and monolingual words, especially when dealing with diverse languages. To overcome these limitations, we propose an Embedded Heterogeneous Attention Transformer (EHAT) to establish cross-domain relationships and local correspondences between images and different languages by using a heterogeneous network.  EHAT comprises Masked Heterogeneous Cross-attention (MHCA), Heterogeneous Attention Reasoning Network (HARN), and Heterogeneous Co-attention (HCA). The HARN serves as the core network and it captures cross-domain relationships by leveraging visual bounding box representation features to connect word features from two languages and to learn heterogeneous maps. MHCA and HCA facilitate cross-domain integration in the encoder through specialized heterogeneous attention mechanisms, enabling a single model to generate captions in two languages. We evaluate our approach on the MSCOCO dataset to generate captions in English and Chinese, two languages that exhibit significant differences in their language families. The experimental results demonstrate the superior performance of our method compared to existing advanced monolingual methods. Our proposed EHAT framework effectively addresses the challenges of cross-lingual image captioning, paving the way for improved multilingual image analysis and understanding.

\end{abstract}

\begin{IEEEkeywords}
Image Captioning, Cross-lingual Learning, Cross-model Learning, Heterogeneous Attention Reasoning.
\end{IEEEkeywords}

\section{Introduction}

Image captioning aims to generate language descriptions from image information. As one of the prominent cross-modal tasks, it holds immense potential in multimedia analysis and applications. Owing to the development of deep learning with semantic understanding and alignment between modalities, the focus of multimedia description has shifted from limited word-level descriptions~\cite{chua2009nus, xie2013picture, cui2017general} to more abundant and detailed sentence-level captions~\cite{chen2015microsoft, rennie2017self, pan2020x,yang2020auto,nguyen2022grit}. However, captioning models cannot be confined to generating just one language in our society with globalized communication. Despite the growing interest in cross-lingual image captioning tasks in recent years~\cite{li2019coco, miyazaki2016cross, elliott2015multilingual}, the generation of captions in two or more languages remains a significant challenge.

\begin{figure}[!t]
\centering
\includegraphics[width=3.5in]{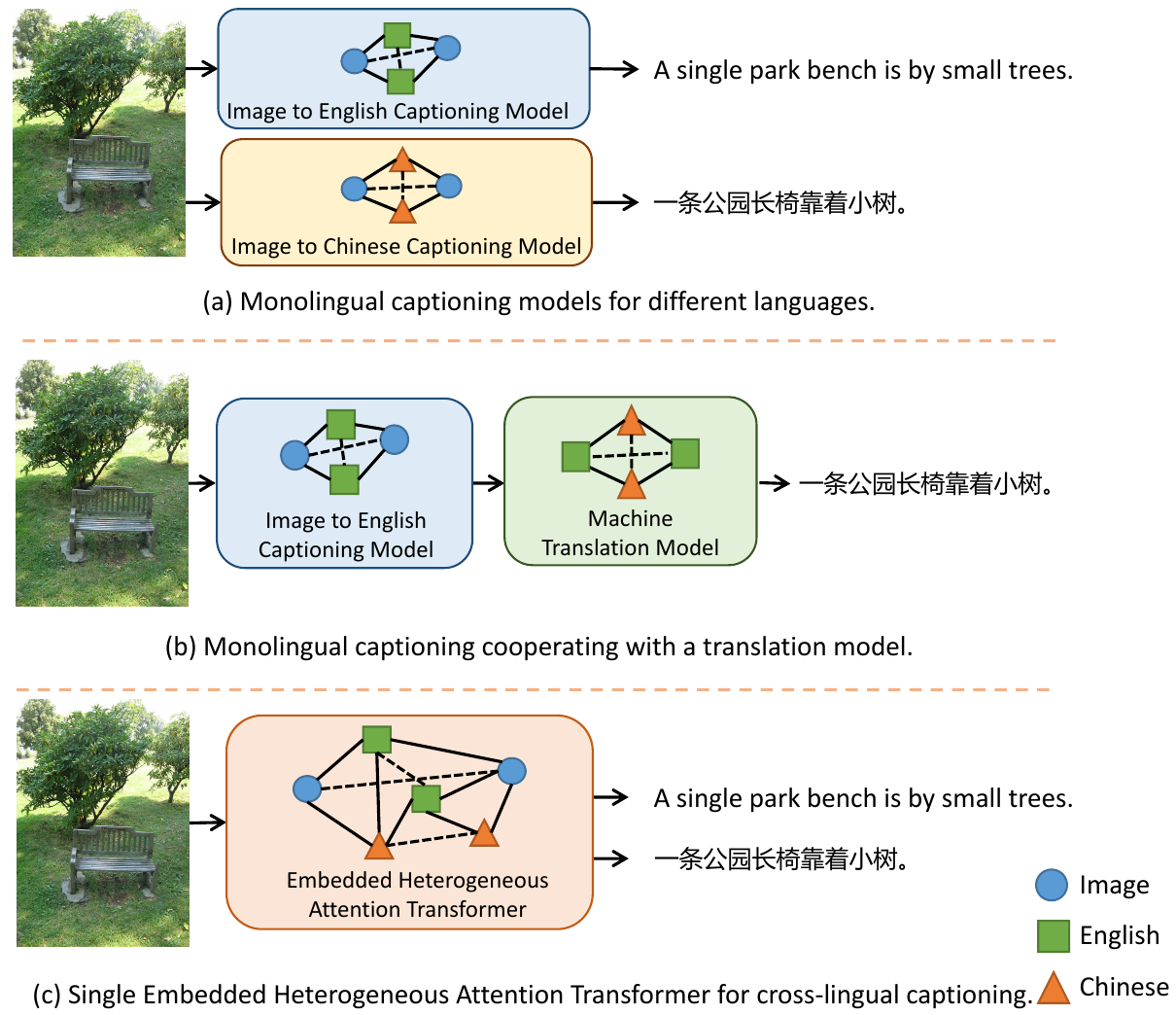}
\caption{Compared with the existing cross-lingual image captioning schemes (a) and (b), our method (c) is designed on the ensemble captioning model to generate multiple languages, which is more compact and efficient.}
\label{methods}
\end{figure}

When dealing with multiple heterogeneous data for cross-modal language generation tasks, first visual information is characterized by high redundancy and sparsity. This occurs because only a small fraction of the regions of interest in an image are meaningful compared with accurate and concise textual modalities. There are also significant differences between different languages. For example, Chinese and English belong to different language families, and they exhibit notable distinctions in terms of sentence structure and semantic coherence. These inherent challenges make it difficult to generate high-quality cross-lingual descriptions simultaneously. The existing cross-lingual image captioning methods~\cite{li2019coco, miyazaki2016cross,elliott2015multilingual,anderson2018bottom,chen2021towards,wang2020cross,huang2019attention,jaffe2017generating,gu2018unpaired} are limited and often rely on multistep implementations. These approaches typically involve training separate generation models using different language datasets~\cite{chen2021towards}, as shown in Fig.~\ref{methods}~(a), or linking captioning models with translation models~\cite{jaffe2017generating,gu2018unpaired}, as shown in Fig.~\ref{methods}~(b). However, these methods require multiple monolingual captioning models to independently manage linguistic interactions in each pathway. Monolingual captioning with a translation model fails to consider the local matching between visual regions and translated caption words, where a false monolingual generated caption directly leads to a false translation result. Both approaches merely mitigate the potential interference caused by semantic differences between the two languages by using a single model without visual guidance, leading to inaccuracies and inconsistencies in the generated captions. In our work, we specifically focus on the cross-lingual image captioning task, which involves simultaneously generating precise and fluent captions in two widely used languages: English and Chinese. As illustrated in Fig.~\ref{methods}~(c), we design a single-ensemble cross-lingual captioning model with heterogeneous modeling to effectively capture the correlation between cross-modal and cross-lingual aspects of this complicated task.

Unlike previous heterogeneous CNN-LSTM encoder-decoder structures~\cite{vinyals2015show, xu2015show, anderson2018bottom, jiang2018recurrent, yao2018exploring}, the transformer~\cite{vaswani2017attention} architecture provides a homogeneous approach by unifying the different modality inputs into semantic embeddings. This develop has produced to breakthrough advancements in cross-modal task models. Numerous Vision-Language Pretraining (VLP) models have been proposed~\cite{lu2019vilbert,li2020oscar,radford2021learning,alayrac2022flamingo}, which leverage large amounts of image-text data to train the encoder and provide improved prior representations for downstream tasks. While the self-attention mechanism captures the interdependencies among cross-modal tokens, it fails to fully exploit the heterogeneous nature of vision and different languages. Specifically, the self-attention mechanism overlooks the structural alignments between image regions and corresponding textual descriptions. Yu \textit{et al.}~\cite{yu2021heterogeneous} introduced heterogeneous attention to cross-modal retrieval task and demonstrated that heterogeneous attention can facilitate more precise matching of different modalities of information. Our work is based on the visual model of VinVL~\cite{zhang2021vinvl}, a transformer-based VLP model that can learn object instance-centered relationships between visual and language domains. We enhance the downstream captioning task and introduce heterogeneous attention mechanisms to facilitate the generation of multilingual captions only in the transformer decoder.

Our objective is to design a transformer-based cross-modal and cross-lingual alignment module for efficient cross-lingual image captioning. This module not only connects three heterogeneous data types, image, English and Chinese, but also places special emphasis on local matching through joint embedding within the transformer-based training framework. In particular, this is achieved by heterogeneous attention with anchored visual input to model local-level heterogeneous relationships embedded into the transformer structure. We demonstrate that such heterogeneous attention is instrumental in generating precise and fluent English-Chinese captions simultaneously from visual information. To accomplish this goal and to incorporate cross-lingual parallel interaction, we propose Embedded Heterogeneous Attention Transformer (EHAT) model in Transformer decoder, composed of three components orderly: Masked Heterogeneous Cross-attention (MHCA), Heterogeneous Attention Reasoning Network (HARN) and Heterogeneous Co-attention (HCA). Initially, MHCA, which serves as the initial stage of the model with the mask function and self-attention, aligns the dimensional space between image region features and language embeddings. Second, HARN, acting as the core of EHAT, aligns semantic information between each image-language pair through cross-attention, and it provides heterogeneous similarity weights to establish correlations between English and Chinese, anchored by the visual context. Finally, HCA handles the final language representations sent to the generator and facilitates cross-lingual interaction. In addition, we design two variants of HARN to explore the intricate mutual interference arising from heterogeneous associations. Notably, our method represents the first application of heterogeneous attention embedded into a transformer decoder for cross-lingual image captioning in a single ensemble structure, effectively capturing both global and local features.

The main contributions of this paper are summarized as follows:

\begin{itemize}
\item We propose the Embedded Heterogeneous Attention Transformer (EHAT) for cross-lingual image captioning, which leverages the benefits of large batch computations and global matching while incorporating local attention and cross-domain semantic alignment capabilities from heterogeneous data.

\item We present the Heterogeneous Attention Reasoning Network (HARN) to establish heterogeneous attention alignment with the visual modality as the anchor, facilitating the relationship structure between two languages. It effectively bridges the local matching between visual and textual features, while simultaneously enabling parallel alignment between words in different languages.

\item 
We investigate the effects of language interaction by modifying the heterogeneous structure within the reasoning path. The experiments demonstrate that the effectiveness of incorporating heterogeneous data within a unified model, demonstrating its potential for enhancement.

\end{itemize}

The remainder of this paper is arranged as follows. In Section~\ref{sec2}, we review related work on multimodal vision and language tasks and heterogeneous information learning. In Section~\ref{sec3}, we introduce our proposed method named EHAT and its variants. Section~\ref{sec4} presents the experimental details, results and discussion. Section~\ref{sec5} contains the concluding remarks.

\section{Related work}
\label{sec2}
\subsection{Multimodal Vision and Language Tasks}
Researches~\cite{rennie2017self, yang2018multitask, anderson2018bottom, tan2019comic, xiao2019deep, zhou2020more, zhang2020integrating, yang2020captionnet, guo2020normalized, zhou2021semi, yang2021deconfounded} on image captioning has made great progress and recently the ability to caption images has been taken to the new heights~\cite{huang2021image, zhang2021exploring, yu2021dual, wang2022end, zhou2022compact, wang2022text}. Guo \textit{et al.}~\cite{guo2020normalized} introduced and combined the normalization and geometry-aware methods based on self-attention to construct an image captioning model. Yang \textit{et al.}~\cite{yang2021deconfounded} proposed a deconfounded image captioning framework from a causal perspective to alleviate dataset bias in image captioning. Huang \textit{et al.}~\cite{huang2021image} used the relational context between words and phrases to explore related visual relationships between different objects via a bilinear self-attention model and they encoded more discriminative features in the linguistic context. Wang \textit{et al.}~\cite{wang2022end} proposed an end-to-end transformer-based method focused on the interaction between multiple modalities with a Swin Transformer~\cite{liu2021swin} as visual feature extractor. However, monolingual models can no longer satisfy the requirements for multi-language application scenarios. Growing cross-lingual studies are not limited to one language application in multimodal tasks, such as image captioning~\cite{miyazaki2016cross, lan2017fluency, wang2020cross, gao2022unison}, visual grounding~\cite{dong2020cross}, text image recognition~\cite{chen2022cross}, and image retrieval and tagging~\cite{li2019coco, aggarwal2021towards}. There are still some restrictions for cross-lingual tasks where cross-lingual image captioning is no exception. 

First, existing large-scale image captioning datasets~\cite{chen2015microsoft, plummer2015flickr30k} are mostly monolingual. Commonly used cross-lingual datasets~\cite{li2016adding, lan2017fluency, li2019coco} are relatively insufficient, which may not be conducive for further improvement and generalization of the model. Many lots of methods~\cite{miyazaki2016cross, tsutsui2017using, jia2020icap} cannot reach the same level as monolingual models~\cite{stefanini2021show} in the case of different amounts of data.

Second, the high redundancy of visual information as the only input makes it difficult to select features in image captioning tasks. Chen \textit{et al.}~\cite{chen2021towards} measured the distance of semantic relevance between captions and references, and they computed the cross-model relevance between generated captions and image features for annotation-free evaluation. Gao \textit{et al.}~\cite{gao2022unison} proposed a cross-lingual autoencoder model with sentence parallel corpus mapping and unsupervised feature mapping, that encodes scene graph features from image modality to language. It is more difficult to output two language types of precise and fluent captioning because of a lack of data labels~\cite{miyazaki2016cross,tsutsui2017using,jia2020icap}.

Third, the issue of multilingual interaction in a single model is considered negative~\cite{he2021synchronous}. Wang \textit{et al.}~\cite{wang2019synchronously} proposed an interactive decoder that utilizes paired attention synchronously to generate two languages in neural machine translation. Zhou \textit{et al.}~\cite{zhou2021uc2} proposed a universal cross-lingual cross-modal vision-and-language pretraining method for multilingual image-text retrieval and multilingual visual question answering that implements language transformation via machine translation. At present, research exploring the interactions between languages is still lacking for cross-lingual image captioning.

\subsection{Heterogeneous Information Learning}
Heterogeneous information learning theory~\cite{sun2011pathsim, sun2012mining} has been adopted to model complicated relationships for widely used real-world systems, such as recommendation~\cite{shi2018heterogeneous, hu2018leveraging}, node classification~\cite{fu2017hin2vec, hamilton2017inductive, zhao2021heterogeneous} and text analysis~\cite{linmei2019heterogeneous, hu2020graph}. Wang \textit{et al.}~\cite{wang2022survey} provides a survey where methods and techniques were exhaustively introduced mainly in natural language processing. Considering the increase in the amount of data available for heterogeneous modeling, the complexity also increases and large batches of fast computing methods are needed. Hu \textit{et al.}~\cite{hu2020heterogeneous} proposed a heterogeneous graph transformer architecture for modeling web-scale heterogeneous graphs. A heterogeneous mini-batch graph sampling algorithm for efficient and scalable training has been adopted to motivate many subsequent studies~\cite{yao2020heterogeneous, mei2022relation}.
 
With the progress and improvement of heterogeneous learning research, heterogeneous learning has exhibited powerful relational modeling capabilities and interpretability. This method has been introduced among various multimodal tasks to model semantic alignment between modalities and to build inference paths, such as visual question answering (VQA)~\cite{zhu2020mucko, yu2020cross}, visual commonsense reasoning (VCR)~\cite{yu2019heterogeneous, song2023efficient}, video question answering (VideoQA)~\cite{fan2019heterogeneous, jiang2020reasoning}, video recommendation~\cite{cai2021heterogeneous, cai2022heterogeneous} and cross-modal retrieval~\cite{yu2021heterogeneous}. According to differences among multimodal tasks with inputs, outputs and modal categories, heterogeneous graph modeling and reasoning paths also vary widely. Yu \textit{et al.}~\cite{yu2019heterogeneous} integrated intra-graph and inter-graph reasoning in order to bridge cross domains for VCR. This heterogeneous graph guides answer representations with visual features and question queries to connect different semantic nodes. In the cross-modal retrieval task, Yu \textit{et al.}~\cite{yu2021heterogeneous} presented a heterogeneous attention module to produce the local matching between regions of the image and words in the text, which provides the cross-modal context to learn the relevance. Zhu \textit{et al.}~\cite{zhu2022latent} proposed a novel unified latent heterogeneous representation to explore the complex relationships among samples for incomplete multiview learning. In this work, we first attempt to design a heterogeneous attention learning mechanism for cross-lingual image captioning with alignment in modalities and interaction across languages.

\begin{figure*}[!t]
\centering
\includegraphics[width=6in]{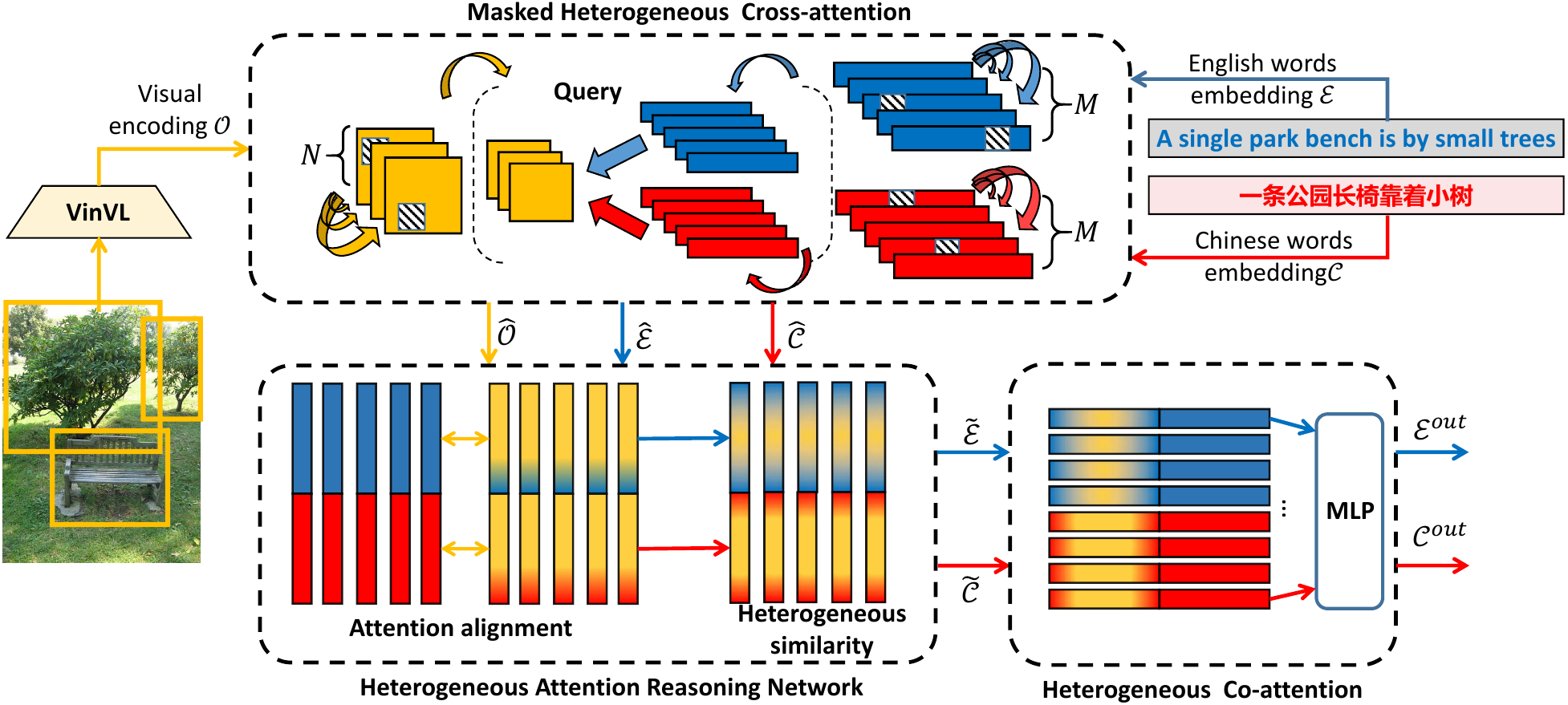}
\caption{The overview of our method. The proposed Embedded Heterogeneous Attention Transformer (EHAT) model is composed of Masked Heterogeneous Cross-attention (MHCA), Heterogeneous Attention Reasoning Network (HARN) and Heterogeneous Co-attention (HCA). MHCA provides dimension space alignment via the mask function and self-attention to handle both image and language. HARN, as the core of EHAT, provides cross-modal semantic alignment via heterogeneous modeling. HCA is focused on cross-lingual interactions and is associated with subsequent caption generation. More specific details and functions of network implementations are discussed in Section~\ref{sec3}. The core module HARN and its variants are shown in Fig.~\ref{variants}. Best view in color.}
\label{framework}
\end{figure*}

\begin{table}[!]
\caption{Explanation of main symbols with their meanings and types.}
\label{tab:Symbol}
\centering
\renewcommand{\arraystretch}{1.2}
\begin{tabular}{ccc}
\hline
Symbol & Meaning & Type \\
\hline
$\mathcal{O}$  &Visual objects encoding  &Tensor matrix  \\
$o_i$   & One of objects encoding  &Tensor vector  \\
$N$    &Number of objects  &Integer  \\
$\mathcal{E},\mathcal{C}$     &EN \& CN caption embeddings &Tensor matrix  \\
$e_i,c_i$     &EN \& CN word embeddings &Tensor vector  \\
$\mathcal{L}$    &Concat of $\mathcal{E}$ and $\mathcal{C}$  &Tensor matrix \\
$M$     &Length of sentence  &Integer  \\
$Q,K,V$   &Query, key and value &Tensor matrix  \\
$d_{model},d_k$   &Dimension of tensor &Integer  \\
$\mathcal{W}$     &Trainable weights  &function  \\
$Q^{\mathcal{O}},K^{\mathcal{O}},V^{\mathcal{O}}$   &Q, K \& V copied from $\mathcal{O}$  &Tensor matrix  \\
$Q^{\mathcal{L}},K^{\mathcal{L}},V^{\mathcal{L}}$   & Q, K \& V copied from $\mathcal{L}$   &Tensor matrix \\
$\hat{\mathcal{O}}, \hat{\mathcal{E}}, \hat{\mathcal{C}}$   &Outputs from MHAC  &Tensor matrix  \\
$\tilde{\mathcal{O}}, \tilde{\mathcal{E}}, \tilde{\mathcal{C}}$   &Outputs from HARN  &Tensor matrix  \\
$\mathcal{E}^{out}, \mathcal{C}^{out}$   &Outputs from HCA  &Tensor matrix  \\
$\Gamma$     &MLP  &function      \\
$\omega^{\mathcal{E}}$, $\omega^{\mathcal{C}}$ &Heterogeneous similarity weight  &Tensor vector\\
$v_1, v_2$ &Subscript for HARN variants & NULL \\
$L_{en}^{CE}(\theta),L_{cn}^{CE}(\theta)$ & Cross-entropy loss & NULL\\
$L_{en}^{RL}(\theta),L_{cn}^{RL}(\theta)$ & Reinforcement learning loss & NULL\\
\hline
\end{tabular}
\end{table}

\section{Method}
\label{sec3}
In this section, we introduce the proposed Embedded Heterogeneous Attention Transformer (EHAT), which is composed of three components: Masked Heterogeneous Cross-attention (MHCA), Heterogeneous Attention Reasoning Network (HARN) and Heterogeneous Co-attention (HCA). To facilitate understanding and to distinguish symbols in our method, we list their meanings and types in Table~\ref{tab:Symbol}.

\subsection{Overview}
Given an image $\mathcal{I}$ paired with different language captions $\mathcal{E}$ (the English caption) and $\mathcal{C}$ (the Chinese caption), we leverage pretrained visual and language encoders and integrate their respective features into our model. The objective is to generate different language captions, i.e., English and Chinese captions, simultaneously.

The MHCA module employs the mask function and the self-attention mechanisms to align the dimensional space between image region features and language embeddings. Then, HARN aligns semantic information across each image-language pair by using cross-attention. It further provides heterogeneous similarity weights to establish correlations between English and Chinese, anchored by the visual context. HARN is the core component of MHCA. Finally, the HCA module processes the final language representations, which are then passed on to the caption generator. HCA facilitates cross-lingual interactions. An overview of our framework is shown in Fig.~\ref{framework}. Based on the VLP encoder, our method is embedded into the decoder of the transformer~\cite{vaswani2017attention} structure.

\subsection{Visual encoding}
We first encode the image into visual features. Each image is represented by a set of object features ${\mathcal{O} = \{o_i\}_{i=1}^N}, o_i \in \mathbb R^{d_k}$ extracted by VinVL~\cite{zhang2021vinvl}, a transformer-based VLP model that can learn object instance-centered relationships between visual and language domains. $N$ is the number of objects in one image. We directly utilize the pretrained model VinVL as our feature extractor, as its encoder component aligns with VinVL. 
Similar to the Transformer encoder~\cite{vaswani2017attention}, our self-attention and cross-attention before the EHAT are built on scaled dot-product attention and multihead Attention, which operate on a query $Q$, key $K$ and value $V$ as:
\begin{align}
\label{atten}
    Attention(Q,K,V) = softmax(\frac{QK^T}{\sqrt{d_k}})V,
\end{align}
where $d_k$ is set as the same dimension for $Q$, $K$ and $V$. Then, multihead attention concatenates the results and projects them with another learned linear projection:
\begin{align}
    H_i = Attention(Q\mathcal{W}^Q,K\mathcal{W}^K,V\mathcal{W}^V), &  \\
    MultiHead(Q,K,V) = Concat(H_1,...H_h) & \mathcal{W}^\mathcal{O},
\end{align}
where trainable weights $\mathcal{W}^Q$, $\mathcal{W}^K$, $\mathcal{W}^V$ $\in \mathbb{R}^{d_{model} \times d_k}$ and $\mathcal{W}^\mathcal{O}$ $\in \mathbb{R}^{d_{model} \times d_k}$. The following is the positionwise feed-forward sublayer (FFN) with max function and linear learning parameters:
\begin{align}
    FFN(x) = max(0, x\mathcal{W}_1 + b_1)\mathcal{W}_2 +b_2.
\end{align}

\subsection{Masked Heterogeneous Cross-attention (MHCA)}

We implement the proposed MHCA module in the transformer decoder layers. 
For the textual part, all words are embedded into the same dimension vectors. The two language embedding representation sets are $\mathcal{E} = \{e_i\}_{i=1}^M, e_i \in \mathbb R^{d_k}$ and $\mathcal{C} = \{c_i\}_{i=1}^M, c_i \in \mathbb R^{d_k}$. Captions of the same image may vary in terms of the number of words across different languages. To address this discrepancy, we unify the word count by considering the maximum length $M$ within a sentence. Both English and Chinese vocabulary are collected on dictionary without low frequency words.

We splice $\mathcal{E}$ and $\mathcal{C}$ together to $\mathcal{L}=[\mathcal{E}, \mathcal{C}], \mathcal{L} \in \mathbb{R}^{2M \times d_k}$ for large batch training. Self-attention performs attention over language embeddings $\mathcal{L}$, i.e., ($Q^\mathcal{L}$ = $K^\mathcal{L}$ = $V^\mathcal{L}$ = $\mathcal{L}$) and cross-attention connects language embeddings $L$ with region features $\mathcal{O}$, i.e., ($Q^\mathcal{L}$ = $\mathcal{L}$, $K^\mathcal{O}$ = $V^\mathcal{O}$ = ${\mathcal{O}}$) as shown in eq.(\ref{atten}):
\begin{align}
    Self(Q^\mathcal{L},K^\mathcal{L},V^\mathcal{L}) = softmax(\frac{Q^\mathcal{L}(K^\mathcal{L})^T}{\sqrt{d_k}})V^\mathcal{L},  \\
    Cross(Q^\mathcal{L},K^\mathcal{O},V^\mathcal{O}) = softmax(\frac{Q^\mathcal{L}(K^\mathcal{O})^T}{\sqrt{d_k}})V^\mathcal{O},
\end{align}

The central objective of EHAT is to construct heterogeneous relationships and spatial alignment, which requires the heterogeneous attention module to have aligned vector dimensions for the nodes. However, visual features and language embeddings often differ significantly in their dimensionalities. To address this challenge, we introduce the Masked Heterogeneous Cross-attention (MHCA) mechanism. MHCA is specifically designed to establish a masked attention mechanism for handling heterogeneous data and for unifying the vector dimensions during the attention calculation process. By incorporating MHCA, EHAT effectively addresses the dimension disparity between visual features and language embeddings, enabling seamless integration and alignment within the heterogeneous attention framework.

Specific to a single image, given $N$ visual features ${\mathcal{O} \in \mathbb{R}^{N \times d_k}}$ and $M$ bilingual captioning features ${\mathcal{E} \in \mathbb{R}^{M \times d_k}}$, ${\mathcal{C} \in \mathbb{R}^{M \times d_k}}$, MHCA outputs $\hat{\mathcal{O}}, \hat{\mathcal{E}}, \hat{\mathcal{C}}$ as the space aligned representations of visual objects and text words. In datasets, different samples often have varying numbers of bounding boxes and caption lengths within their respective feature spaces. For heterogeneous embedding, the unified dimension space for all heterogeneous nodes should be ensured to compute heterogeneous similarity and embed into decoder. However, the number of bounding boxes $N$ and the length of captions $M$ frequently differ and can vary depending on the images. Simply  cropping $N$ or expanding $M$ for alignment is not reasonable, as it may introduce captioning bias without a clear rationale or intuitive judgment. Thus, we design MHCA inspired by the self-attention mechanism and outputs as follows:
\begin{align}
  \hat{\mathcal{O}}, \hat{\mathcal{E}},\hat{\mathcal{C}} = MHCA(\mathcal{O},\mathcal{E}, \mathcal{C}).
\end{align}

The specific transformation process is the same for the three inputs. We take $\mathcal{O}$ as an example to list the calculation formula:
\begin{align}
  \mathcal{O}^{att} &= softmax(\frac{\mathcal{O}(\mathcal{O})^T}{\sqrt{d_k}}),   \\
  \mathcal{O}^{mask} =   \; &dropout(mask[\mathcal{O}^{att}])\mathcal{O},  \\
  \hat{\mathcal{O}} = \;  d&ropout(\frac{(\mathcal{O}^{mask})^T\mathcal{O}^{mask}}{\sqrt{d_k}}),
\end{align}
$\hat{\mathcal{E}}$, $\hat{\mathcal{C}}$ can be obtained in the same way from $\mathcal{E}$, $\mathcal{C}$ and all of them are mapped into $\mathbb{R}^{d_k \times d_k}$. The deep transformer structure and complex heterogeneous attention can easily result in network redundancy and can lead to training overfitting from multiple matrix calculations. Masking and two dropout operations can relieve this misgiving at the network level. The mask function and dropout are the same as the settings in the self-attention mechanism\cite{vaswani2017attention}. MHCA provides favorable alignment conditions for heterogeneous alignment construction and linear operations in reasoning.

\subsection{Heterogeneous Attention Reasoning Network (HARN)}
HARN is designed to establish the local matching between visual objects and language words. Triples $[\hat{\mathcal{O}}, \hat{\mathcal{E}}, \hat{\mathcal{C}}]$ are used to construct heterogeneous relationships. Our initial setting as shown in Fig.~\ref{variants}(a), is to avoid the two languages influencing each other when building the maps. Vision features are provided as anchors to compute attention from both languages. We define the HARN module as follows:
\begin{align}
\label{HARN}
  \tilde{\mathcal{E}}, \tilde{\mathcal{C}} = HARN(\hat{\mathcal{O}}, \hat{\mathcal{E}}, \hat{\mathcal{C}}).
\end{align}
Visual features are anchored to generate two directed links to language embeddings by two MLPs, which are set to two queries $\hat{\mathcal{O}}_1 = \hat{\mathcal{O}}_2 = \hat{\mathcal{O}}$. This means that $\hat{\mathcal{E}}, \hat{\mathcal{C}}$ have their own respective attention maps to obtain an enhanced language representation $\tilde{\mathcal{E}}, \tilde{\mathcal{C}}$ by anchored local visual features with heterogeneous similarity calculation. The specific calculation processes, where reasoning for alignment is calculated by the heterogeneous similarity, are divided into two paths:
\begin{align}
\label{path1}
  \tilde{\mathcal{E}} = \Gamma^H_{\mathcal{E}}([1, \omega^{\mathcal{E}}]\hat{\mathcal{E}}),  \\
\label{path2}
  \tilde{\mathcal{C}} = \Gamma^H_{\mathcal{C}}([1, \omega^{\mathcal{C}}]\hat{\mathcal{C}}),
\end{align}
where $\Gamma^H_{\mathcal{E}}, \Gamma^H_{\mathcal{C}}$ indicate different MLP networks and $\omega^{\mathcal{E}}$, $\omega^{\mathcal{C}}$ represent heterogeneous similarity weights. $[., .]$ indicates the concatenation operation. The feature dimensions are unified to $d_k$ by MLP networks. For simplicity we use the fully connected layer (FC) as the MLP. We implement the heterogeneous similarity weight by $\omega^{\mathcal{E}}$ and $\omega^{\mathcal{C}}$, which can be calculated as:
\begin{align}
  \omega^{\mathcal{E}} = \frac{exp(\Gamma_{\mathcal{E}}([\hat{\mathcal{E}}, \hat{\mathcal{O}}_1])\mathcal{W}_{\mathcal{E}})}{exp(\Gamma_{\mathcal{E}}([\hat{\mathcal{E}}, \hat{\mathcal{O}}_1])\mathcal{W}_{\mathcal{E}}) + exp(\Gamma_{\mathcal{O}_1}(\hat{\mathcal{O}}_1)\mathcal{W}_{\mathcal{O}_1})},   \\
  \omega^{\mathcal{C}} = \frac{exp(\Gamma_{\mathcal{C}}([\hat{\mathcal{C}}, \hat{\mathcal{O}}_2])\mathcal{W}_{\mathcal{C}})}{exp(\Gamma_{\mathcal{C}}([\hat{\mathcal{C}}, \hat{\mathcal{O}}_2])\mathcal{W}_{\mathcal{C}}) + exp(\Gamma_{\mathcal{O}_2}(\hat{\mathcal{O}}_2)\mathcal{W}_{\mathcal{O}_2})},  
\end{align}
where $\mathcal{W}_{\mathcal{E}}, \mathcal{W}_{\mathcal{O}_1}, \mathcal{W}_{\mathcal{C}}, \mathcal{W}_{\mathcal{O}_2}$ are trainable weighted parameters, and $\Gamma_{\mathcal{E}}, \Gamma_{\mathcal{C}}$ are similar to $\Gamma^H_{\mathcal{E}}, \Gamma^H_{\mathcal{C}}$. $\Gamma_{\mathcal{O}_1}, \Gamma_{\mathcal{O}_2}$ are designed as:
\begin{align}
  \Gamma_{\mathcal{O}_j} = softmax(FC(\hat{\mathcal{O}}_j))\hat{\mathcal{O}}_j, j=1,2.
\end{align}

The algorithm here increases the visual attention information of the language and fuses into itself separately. Although $\hat{\mathcal{E}}, \hat{\mathcal{C}}$ search for interest attention areas of interest to form linguistically specific captions without mutual interference, anchor homology makes bilingual words closely related which is demonstrated by attention visualization in Section~\ref{sec4}.C.

\subsection{HARN Variants}
\begin{figure*}[!t]
\centering
\includegraphics[width=5.7in]{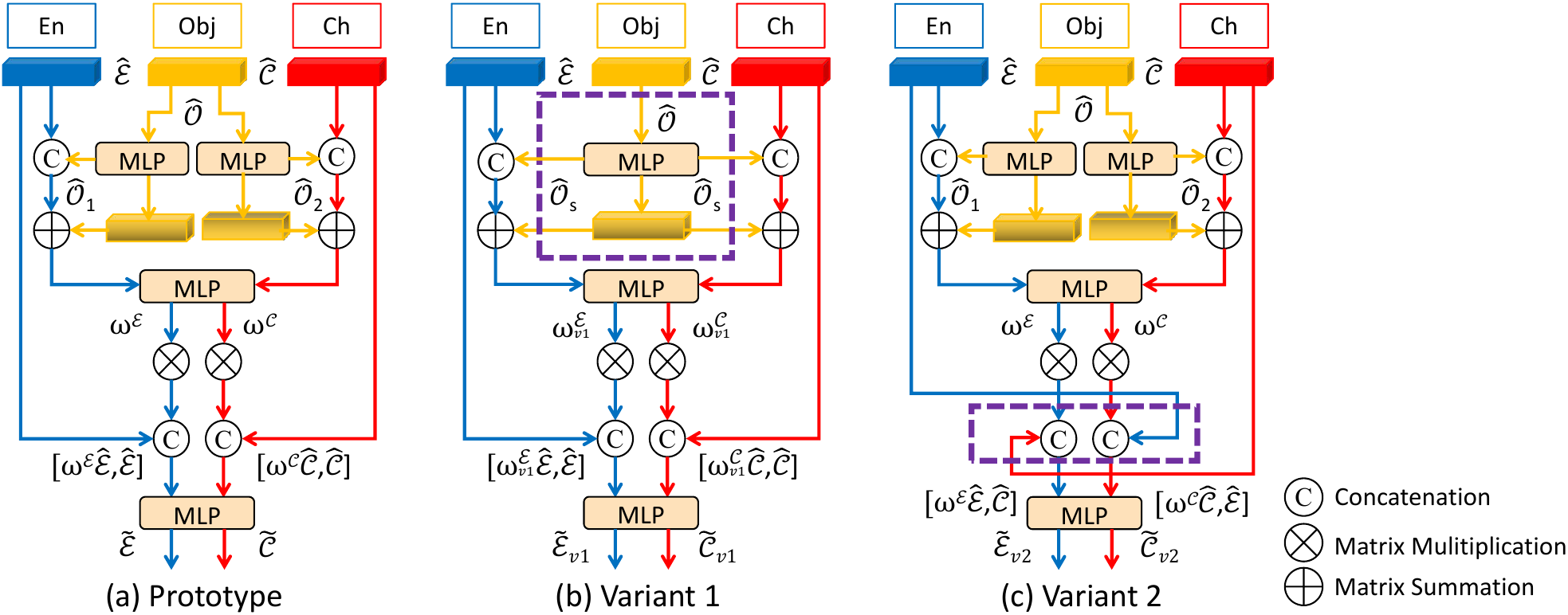}
\caption{There are two variants for HARN (b) and (c) compared with prototype one (a). The MLP includes a combination of different operations such as FC and softmax. The change in Variant~1 is adjusted with anchored visual features via attention. The change in Variant~2 is the direct transposition of linguistic connection to increase interaction. The changes are marked in purple.}
\label{variants}
\end{figure*}

To explore the potential impact of language interactions on heterogeneous attention structures, we attempt to change the computational mode to the path of reasoning, which can intuitively establish a direct correlation between the two languages. Note that heterogeneous attention structure is not unique, and inference paths are also diverse. We list two types of variants that may be persuasive as shown in Fig.~\ref{variants}~(b) and Fig.~\ref{variants}~(c).

According to the anchored visual features, our Variant~1 limits two languages that are focused on the same attention maps. English and Chinese have large gaps in terms of vocabulary, grammar, and even expression. However, it is difficult to verify whether the interest points of the image are also different for machine captioning. Thus, we change the queries $\hat{\mathcal{O}}_1 = \hat{\mathcal{O}}_2 = \hat{\mathcal{O}}$ to $\hat{\mathcal{O}}_s = \hat{\mathcal{O}}$ through only one MLP. The cross-attention on the visual part can be forced to unify for two languages as shown in Fig.~\ref{variants}~(b). The inputs and outputs should be changed from $eq.$(\ref{HARN}) to below:
\begin{align}
  \tilde{\mathcal{E}}_{v1}, \tilde{\mathcal{C}}_{v1} = HARN(\hat{\mathcal{O}}, \hat{\mathcal{E}}, \hat{\mathcal{C}}),
\end{align}
where $\hat{\mathcal{O}}_s$ is the module input information for vision. In addition, heterogeneous attention reasoning should be changed by $\omega^{\mathcal{E}}_{v1}$ and $\omega^{\mathcal{C}}_{v1}$:
\begin{align}
  \omega^{\mathcal{E}}_{v1} = \frac{exp(\Gamma_{\mathcal{E}}([\hat{\mathcal{E}}, \hat{\mathcal{O}}_s])\mathcal{W}_{\mathcal{E}})}{exp(\Gamma_{\mathcal{E}}([\hat{\mathcal{E}}, \hat{\mathcal{O}}_s])\mathcal{W}_{\mathcal{E}}) + exp(\Gamma_{\mathcal{O}_s}(\hat{\mathcal{O}}_s)\mathcal{W}_{\mathcal{O}_s})},   \\
  \omega^{\mathcal{C}}_{v1} = \frac{exp(\Gamma_{\mathcal{C}}([\hat{\mathcal{C}}, \hat{\mathcal{O}}_s])\mathcal{W}_{\mathcal{C}})}{exp(\Gamma_{\mathcal{C}}([\hat{\mathcal{C}}, \hat{\mathcal{O}}_s])\mathcal{W}_{\mathcal{C}}) + exp(\Gamma_{\mathcal{O}_s}(\hat{\mathcal{O}}_s)\mathcal{W}_{\mathcal{O}_s})}. 
\end{align}
Moreover, Variant~1 reduces the half parameters for visual attention and fusion resulted in lightweight.

Variant~2 as shown in Fig.~\ref{variants}~(c) is more intuitive for increasing cross-lingual interactions. We change two paths $eq.$(\ref{path1}), $eq.$(\ref{path2}) and respectively combine them with another language as a reference to push interaction as follows:
\begin{align}
  \tilde{\mathcal{E}}_{v2} = \Gamma^H_{\mathcal{E}}([\omega^{\mathcal{E}}\hat{\mathcal{E}}, \hat{\mathcal{C}}]),   \\
  \tilde{\mathcal{C}}_{v2} = \Gamma^H_{\mathcal{C}}([\omega^{\mathcal{C}}\hat{\mathcal{C}}, \hat{\mathcal{E}}]).
\end{align}
We implement this variant to verify that either a positive or negative effect can be found by comparison with a prototype for mutual influence in a single ensemble cross-lingual image captioning.

\subsection{Heterogeneous Co-attention (HCA)}
HCA, as the end of EHAT, produces $\tilde{\mathcal{E}}, \tilde{\mathcal{C}} \in \mathbb{R}^{d_k \times d_k}$ which can be called heterogeneous co-attention to manage the final language representation. This module introduces heterogeneous information into language embeddings to generate captions in pairs.
\begin{align}
  \mathcal{E}^{out} = (1 +\lambda softmax(\frac{\mathcal{E} \tilde{\mathcal{E}}}{\sqrt{d_k}}))\mathcal{E}, \\
  \mathcal{C}^{out} = (1 +\lambda softmax(\frac{\mathcal{C} \tilde{\mathcal{C}}}{\sqrt{d_k}}))\mathcal{C},
\end{align}
where $\lambda$ is a shared heterogeneous hyperparameter and $ \mathcal{E}^{out}, \mathcal{C}^{out} \in \mathbb{R}^{M \times d_k}$ are simultaneously sent to the generator. As mentioned in Section~\ref{sec2}.A and as inspired by recent work\cite{he2021synchronous}, $\lambda$ does not have an experience value that must be selected by experiments. Intuitively, it can be affected by the dataset and training model. Our test results are shown in Table~\ref{tab:lambda}.

\subsection{Loss Functions}
To train our image captioning model with EHAT, two-stage optimization is adopted as in \cite{yao2018exploring, rennie2017self, gu2018stack, zhang2021exploring}. Here, we use $ \{ \hat{\mathcal{E}}_{t} \}^{T}_{t=1} $ and $ \{ \hat{\mathcal{C}}_{t} \}^{T}_{t=1} $ to represent for English and Chinese sentence words, respectively. $t$ represents the word under the current time-step and $T$ is the set of words in a caption sentence. In the first stage, we optimize the model by minimizing the cross-entropy loss $L^{CE}(\theta)$ of the two output language sentences, where the decoder predicts each token in parallel and maintains interactions with each other:
\begin{align}
L_{en}^{CE}(\theta) & = -\sum_{t=1}^{T}log(p_{\theta}(\hat{\mathcal{E}}_{t}|\hat{\mathcal{E}}_{1:t-1},\hat{\mathcal{C}}_{1:t-1})), \\
L_{cn}^{CE}(\theta) & = -\sum_{t=1}^{T}log(p_{\theta}(\hat{\mathcal{C}}_{t}|\hat{\mathcal{E}}_{1:t-1},\hat{\mathcal{C}}_{1:t-1})), \\
& L^{CE}(\theta) =  L_{en}^{CE}(\theta) + L_{cn}^{CE}(\theta),
\end{align}
where $\hat{\mathcal{E}}_{1:t-1}$ and $\hat{\mathcal{C}}_{1:t-1}$ are the ground truth words listed from 1 to t-1 in the caption sentences. The $t$-th word is predicted according to 1:t-1 for both $\hat{\mathcal{E}}_t$ and $\hat{\mathcal{C}}_t$. $\theta$ represents all the model parameters in the first stage.

In the second training stage, we directly optimize our model with the CIDEr score\cite{luo2020better} as a reinforcement learning training metric based on the best model trained with cross-entropy loss in the first stage. We minimize the negative expected reward of the sampled sentence as the reinforcement learning loss $L^{RL}(\theta)$:
\begin{align}
L_{en}^{RL}(\theta) & = E_{\tilde{\mathcal{E}}_{1:T}\sim p_{\theta}}[R(\tilde{\mathcal{E}}_{1:T})], \\
L_{cn}^{RL}(\theta) & = E_{\tilde{\mathcal{C}}_{1:T}\sim p_{\theta}}[R(\tilde{\mathcal{C}}_{1:T})], \\
L^{RL}(\theta) & =  L_{en}^{RL}(\theta) + L_{cn}^{RL}(\theta),
\end{align}
where $\tilde{\mathcal{E}}_{1:T}$ and $\tilde{\mathcal{C}}_{1:T}$ are the sampled caption sentences. $R()$ is the CIDEr score function. $E[]$ is the calculation of the expectation from multiple samples. We also adopt the algorithm proposed in self-critical sequence training\cite{rennie2017self}. The loss gradient $\nabla_{\theta}L^{RL}(\theta)$ can be approximated by a single Monte-Carlo sampled sentence as:
\begin{align}
\nabla_{\theta}L_{en}^{RL}(\theta) \approx & \ [L_{en}^{RL}(\theta)-b_{en}]\nabla_{\theta}log(p_{\theta}(\tilde{\mathcal{E}}_{1:T})),\\
\nabla_{\theta}L_{cn}^{RL}(\theta) \approx & \ [L_{cn}^{RL}(\theta)-b_{cn}]\nabla_{\theta}log(p_{\theta}(\tilde{\mathcal{C}}_{1:T})),\\
\nabla_{\theta}L^{RL}(\theta) & = \nabla_{\theta}L_{en}^{RL}(\theta) + \nabla_{\theta}L_{cn}^{RL}(\theta),
\end{align}
where $b_{en}$ and $b_{cn}$ are the English and Chinese baseline scores, respectively, which are defined as the average reward of the remaining samples rather than the original greedy decoding reward. $\theta$ still represents all model parameters but in the second stage.


\section{Experiments}
\label{sec4}
\subsection{Dataset and Evaluation Metrics}

\begin{figure*}[!t]
\centering
\includegraphics[width=6.0in]{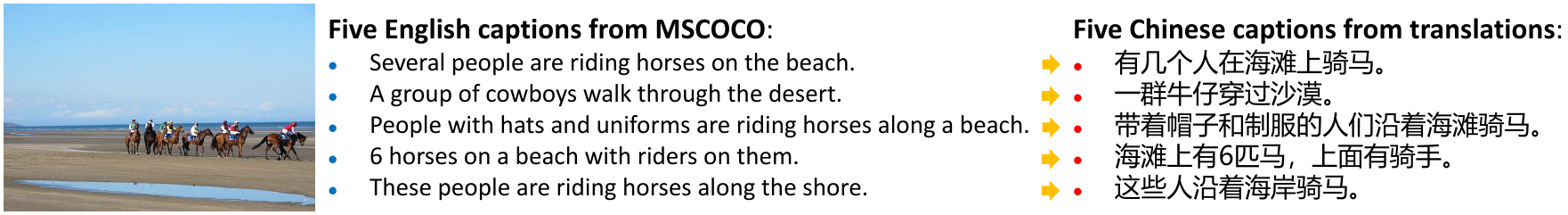}
\caption{A sample of the experimental dataset. The image has five English and five Chinese descriptions. The consistency of the captions is ensured by translation and manual proofreading.}
\label{data}
\end{figure*}

\begin{figure}[!t]
\centering
\includegraphics[width=2.5in]{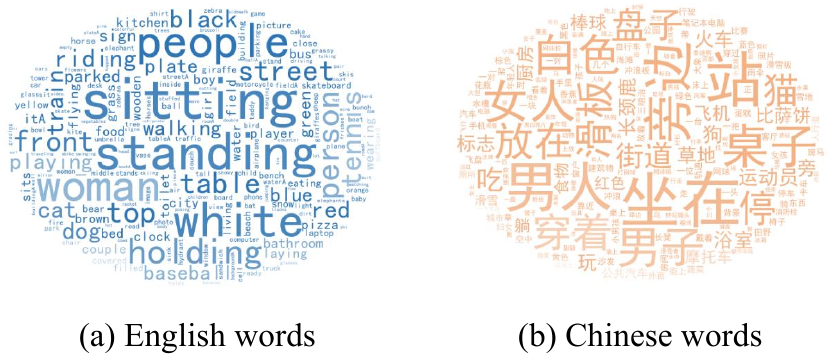}
\caption{Word clouds for the dataset. The font size in the word cloud represents the frequency of word occurrences, with larger font indicating higher frequencies.}
\label{clouds}
\end{figure}

\begin{table}[!t]
\caption{Distribution of top 10 English and Chinese words Frequency in the dataset.}
\label{tab:Topwords}
\centering
\begin{tabular}{cccc}
\hline
English word & Frequency & Chinese word  &Frequency \\
\hline
sitting   &54800  & \rule[-2pt]{0mm}{0.35cm} \multirow{10}{*}{\includegraphics[width=0.03\textwidth,height=0.128\textheight]{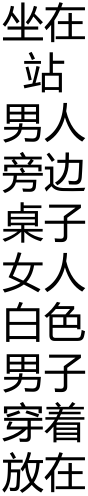}} &45676 \\
standing  &43929  &     &45184\\
white     &36608  &     &35852\\
people    &35518  &     &32383\\
women     &33386  &     &27830\\
holding   &28887  &     &26175\\
person    &24512  &     &24077\\
top       &20956  &     &22481\\
front     &19715  &     &11816\\
table     &19459  &     &17794\\

\hline
\end{tabular}
\end{table}


Our experiments are supported by the whole large-scale MSCOCO dataset~\cite{chen2015microsoft} and we use the Baidu translation API to extend the image captions from English sentences to Chinese, which are also proofread by manual adjustment. To facilitate better management of the Chinese and English captions within the dataset, we performed separate word segmentation processes. For the English captions, we utilized natural word boundaries and constructed an English dictionary by removing words with frequencies lower than five. The final English vocabulary consisted of 9487 words. Similarly, for the Chinese captions, we constructed a Chinese dictionary by removing words with frequencies lower than ten. The final Chinese vocabulary consisted of 9532 words consisted Chinese single words and phrases. After revision and refinement, a total of 123,287 unique images and 1,233,534 high-quality captions were produced. On average, there are five pairs of English and Chinese captions for each image. Fig.~\ref{data} illustrates a sample of the experimental dataset. To visually display the word frequencies in the Chinese and English captions within the dataset, we conducted separate visualizations of the Chinese and English words in Fig.~\ref{clouds}. Moreover, we enumerated the top 10 most frequently occurring English and Chinese words in Table~\ref{tab:Topwords}, along with their respective frequencies. It is worth noting that the corresponding rankings are not entirely identical, which indicates differences in linguistic expression between the two languages.

With advancements in translation technology, manual proofreading is primarily focused on assessing the appropriateness of semantic logic. As a result, the verification process for a single sample takes an average of 3 seconds, leading to a total workload of approximately 500 hours. The experimental results significantly show that more training data is beneficial for improving the fluency and the accuracy of captioning generation.

We follow the 'Karpathy' splits~\cite{karpathy2015deep} where the dataset is divided into 113,287 training images, 5,000 validation images, and 5,000 testing images. We evaluate our model with commonly used image description generation metrics, BLEU-1/4, METEOR, ROUGE, and CIDEr, denoted as B@1/4, M, R, and C respectively.

\subsection{Main Results with Settings}

\begin{table}[!t]
\caption{Performance for generating English captions comparison with the advanced monolingual methods on the COCO 'Karpathy'~\cite{karpathy2015deep} test split with standard metrics. Baseline~(En-only) means proposed model trained with English captions only.}
\label{tab:En}
\centering
\begin{tabular}{c|ccccc}
\hline
Model & B@1 & B@4  & M  &R &C\\
\hline
Up-Down\cite{anderson2018bottom} & 79.8  &36.3  &27.7  &56.9  &120.1\\
RFNet\cite{jiang2018recurrent} & 80.4  &37.9  &28.3  &58.3  &125.7\\
GCN-LSTM\cite{yao2018exploring} & 80.5 & 38.2 & 28.5 & 58.3 & 127.6\\
SGAE\cite{yang2019auto} & 80.8 & 38.4 & 28.4 & 58.6 & 127.8\\
ORT\cite{herdade2019image} & 80.5 & 38.6 &28.7 & 58.4 & 128.3\\
AoANet\cite{huang2019attention} & 80.2 & 38.9 & 29.2 & 58.8 & 129.8\\
$M^2$ Transformer\cite{cornia2020meshed} & 80.8 & 39.1 & 29.2 & 58.6 & 131.2\\
TCIC\cite{fan2021tcic} & 80.9 & 39.7 & 29.2 & 58.6 & 132.9\\
X-Transformer\cite{pan2020x} & 80.9 & 39.7 & 29.5 & 59.1 & 132.8\\
RSTNet\cite{zhang2021rstnet} & 81.1 & 39.3 & 29.4 & 58.8 & 133.3\\
$S^2$ Transformer\cite{zeng2022s2} & 81.1 & 39.6 & \textbf{29.6} & 59.1 & \textbf{133.5}\\
\hline
Baseline~(en-only) & 81.1 & 39.1 & 29.0 & 59.2 & 130.3\\
EHAT & \textbf{81.9} & \textbf{40.1} & \textbf{29.6} & \textbf{59.4} & \textbf{133.5}\\
\hline
\end{tabular}
\end{table}

\begin{table}[!t]
\caption{Performance for generating Chinese captions comparison with the advanced monolingual methods on the COCO 'Karpathy'~\cite{karpathy2015deep} test split with standard metrics. Baseline~(cn-only) means proposed model trained with Chinese captions only.}
\label{tab:Ch}
\centering
\begin{tabular}{c|ccccc}
\hline
Model & B@1 & B@4  & M  &R &C\\
\hline
ORT\cite{herdade2019image} & 77.4 & 31.5 & 30.5 & 54.0 & 107.1\\
AoANet\cite{huang2019attention} & 77.9 & 32.7 & 31.0 & 54.6 & 109.0\\
$M^2$ Transformer\cite{cornia2020meshed} & 78.6 & 32.6 & 31.1 & 54.4 & 109.9\\
\hline
Baseline~(cn-only) & 78.4 & 32.5 & 30.9 & 54.7 & 109.2\\
EHAT & \textbf{79.8} & \textbf{32.9} & \textbf{31.4} & \textbf{55.2} & \textbf{111.6}\\
\hline
\end{tabular}
\end{table}

The settings of the experiments for EHAT are fixed including the prototype and variants. Any changes, such as a shared heterogeneous hyperparameter $\lambda$ are mentioned in subsequent sections. Each image is extracted by VinVL\cite{zhang2021vinvl} and represented as 10 to 50 region features. The English dictionary is collected by more than five times vocabulary with 9,487 words and the Chinese dictionary is collected by more than 10 times vocabulary with 9,532 words. All words are encoded by the torch function nn.Embedding. Captioning length is controlled in 20 words. The number of transformer layers is set to 6. The default hyperparameters of our EHAT are set as follows: the unified attention model dimension and the unified feature dimension $d_{model} = d_k = 512$, all dropout $0.1$, and the shared heterogeneous hyperparameter $\lambda=0.3$. 

In a whole experiment, our training process follows two stages: The first stage trains under cross-entropy loss for 10 epochs, a learning rate with 20,000 warmup steps from 0 to 1e-4 and decreasing to 5e-4. The second stage involves self-critical sequence training~\cite{rennie2017self} with a CIDEr score~\cite{luo2020better} for 15 epochs and the learning rate is 1e-5. Then, every five epochs, the learning rate is reduced $10\%$ from 1e-5 to 1e-7. During training, test results on the validation set are performed every 3K steps, and the model with the best result is saved to the test set for evaluation.

It can be difficult for existing multilingual captioning methods to match the performance of monolingual models due to the lack of studies on cross-lingual image captioning and the inconsistency of the data scale. Thus, we compare our single model multilingual captions with advanced monolingual models separately as shown in Table~\ref{tab:En} and Table~\ref{tab:Ch}. The Chinese caption results in Table~\ref{tab:Ch} are examples of the best models according to our experimental reproductions. Our EHAT captioning metrics FOR both English and Chinese are high. We emphasize that outputting multiple languages is considered difficult and involves mutual interference. However, multilingual collaborative learning essentially increases the abundance of the corpus. Our method tentatively copes with heterogeneous attention modeling on transformer structure and it achieves competitive results against advanced monolingual models. 

Moreover, the ablation study for every component is provided in Table~\ref{tab:component} where results are from the first training stages. Considering that our three components MHCA, HARN and HCA are connected in order with dimension transformation, such that the input dimension is ${\mathcal{E} \in \mathbb{R}^{M \times d_k}}$ but the output dimension is $\hat{\mathcal{E}} \in \mathbb{R}^{d_k \times d_k}$ in MHCA, they cannot be moved directly. Specifically, MHCA and HCA are replaced with matrix multiplication to adjust the dimensions as shown in Line~1 and Line~3. HARN is removed by deleting the heterogeneous similarity weight in Line~2. It is clear that MHCA and HCA, as hubs of feature and dimension transformation, are greatly affected during training. MHCA contributes the defense against overfitting. Removing HCA which is associated with the generator breaks the semantics and coherence of captions. Our model occupies a size of 205 MB and the optimizer occupies 370 MB. Compared with baseline, it is no more than a $9.8\%$ increase for the model and a $10.9\%$ increase for the optimizer. The increase in computational complexity is mostly concentrated in the HARN domain. Line~2 shows that HARN improves performance by establishing a heterogeneous relationship. This approach is acceptable considering the complex computation of heterogeneous similarity to restraint and achieves consistency between Chinese and English captions.

\begin{table*}[!t]
\caption{The ablation study results of three components including MHCA, HARN and HCA for first training stage.}
\label{tab:component}
\centering
\begin{tabular}{c|c|c|c|ccccc|ccccc|c}
\hline
\multicolumn{1}{c|}{\multirow{2}{*}{}} &\multicolumn{1}{c|}{\multirow{2}{*}{MHCA}} &\multicolumn{1}{c|}{\multirow{2}{*}{HARN}} &\multicolumn{1}{c|}{\multirow{2}{*}{HCA}} & \multicolumn{5}{c|}{English} &\multicolumn{5}{c|}{Chinese} &\multicolumn{1}{c}{\multirow{2}{*}{Avg}} \\ 
\cline{5-14}
& & & & B@1 & B@4  & M  &R &C & B@1 & B@4  & M  &R &C \\
\hline
1&        & $\surd$ & $\surd$ &  68.9 &  27.2 & 24.7 & 49.7  & 91.7 & 66.3  & 22.3  & 27.6  &  47.1 & 76.5 & 50.2  \\
2&$\surd$ &         & $\surd$ & 76.1 & 35.1  & 27.5  &  55.2 &  114.0 & 70.5  &  \textbf{27.3} &  27.8 &  48.9 & 89.1 & 57.2 \\
3&$\surd$ & $\surd$ &         & 50.0  & 9.1 & 12.7  &  36.6  & 18.5  &  29.0 & 1.5 & 13.5 & 22.7 & 8.6 &20.2 \\
4&$\surd$ & $\surd$ & $\surd$ & \textbf{77.4} & \textbf{36.1} & \textbf{28.2} & \textbf{56.9} & \textbf{116.1} & \textbf{70.9} & 27.1 & \textbf{28.3} & \textbf{49.8} & \textbf{90.4} & \textbf{58.1}\\
\hline
\end{tabular}
\end{table*}

\subsection{Parameter Analysis and Visualization}
We list the results of the change in the learning rate for the main experiment in Table~\ref{tab:Param}. First, with the help of a reinforcement learning strategy, the model capability is greatly improved from 5e-4 to 1e-5. Then, the performance of the English caption generator increases when the learning rate is reduced $10\%$ from 1e-5 to 1e-7 every 5 epochs. However, the Chinese cations are best at 1e-6 and decrease at 1e-7, which proves that the captioning ability of the model for the two languages is not synchronous. Therefore, this result prompted our subsequent exploration of language interactions.

\begin{table*}[!t]
\caption{The results of the change of the learning rate for main experiment in English and Chinese captions. The 5e-4 is the result of the end for first training stage. 1e-5 to 1e-7 are the results of the change points for second training stage.}
\label{tab:Param}
\centering
\begin{tabular}{c|ccccc|ccccc}
\hline
\multicolumn{1}{c|}{\multirow{2}{*}{EHAT}} & \multicolumn{5}{c|}{English} &\multicolumn{5}{c}{Chinese} \\ 
\cline{2-11}
 & B@1 & B@4  & M  &R &C & B@1 & B@4  & M  &R &C\\
\hline
5e-4 & 77.4 & 36.1 & 28.2 & 56.9 & 116.1 & 70.9 & 27.1 & 28.3 & 49.8 & 90.4\\
1e-5 & 81.4 & 39.8 & 29.3 & 59.5 & 131.9 & 79.5 & 32.9 & 31.4 & 54.9 & 110.6\\
1e-6 & 81.6 & \textbf{40.1} & \textbf{29.6} & \textbf{59.5} & 132.0 & \textbf{79.7} & \textbf{33.4} & \textbf{31.6} & \textbf{55.2} & 111.5\\
1e-7  & \textbf{81.9} & \textbf{40.1} & \textbf{29.6} & 59.4 & \textbf{133.5} & \textbf{79.8} & 32.9 & 31.6 & \textbf{55.2} & \textbf{111.6}\\
\hline
\end{tabular}
\end{table*}

\begin{table*}[!t]
\caption{The results of the change of the shared heterogeneous hyper-parameter $\lambda$ for first training stage in English and Chinese captions. Avg is the average of ten metrics.}
\label{tab:lambda}
\centering
\begin{tabular}{c|ccccc|ccccc|c}
\hline
\multicolumn{1}{c|}{\multirow{2}{*}{$\lambda$}} & \multicolumn{5}{c|}{English} &\multicolumn{5}{c|}{Chinese} &\multicolumn{1}{c}{\multirow{2}{*}{Avg}} \\ 
\cline{2-11}
 & B@1 & B@4  & M  &R &C & B@1 & B@4  & M  &R &C\\
\hline
0.1 & 73.5 & 33.0 & \textbf{28.4} & 55.9 & 109.8 & \textbf{72.2} & 26.8 & 28.7 & 50.0 & 89.6 & 56.8\\
0.3 & \textbf{77.4} & \textbf{36.1} & 28.2 & \textbf{56.9} & \textbf{116.1} & 70.9 & \textbf{27.1} & 28.3 & 49.8 & \textbf{90.4} & \textbf{58.1}\\
0.5 & 71.8 & 30.5 & 27.7 & 54.4 & 106.6 & 71.6 & 26.8 & \textbf{29.0} & \textbf{50.3} & 89.8 & 55.9\\
1.0 & 70.1 & 30.8 & 27.3 &  54.0 & 107.5 & 65.5 & 19.6 & 25.7 & 44.3 & 73.6 & 51.8\\
\hline
\end{tabular}
\end{table*}

The shared heterogeneous hyperparameter $\lambda$ controls the fusion proportion of heterogeneous attention learning features. Although heterogeneous modeling has powerful reasoning and alignment abilities, it is complex, and involves combinations of fully connected layers and linear transformations. Compared with that produced by the transformer structure, the model easily causes overfitting even if dropout layers are frequently added to the heterogeneous structure. We tested the $\lambda$ values listed in Table~\ref{tab:lambda}. The model produces premature overfitting in the first training stage when $\lambda = 1$ and the effect of Chinese captions is worse. It is found that $\lambda \leq 0.5$ can cause the convergence of the model to a controllable value. We calculate the average of the metrics for each $\lambda$ value to choose the best 0.3. This indicates that such a value does not have much influence on the subsequent training when $\lambda$ is less than 0.5. 

\begin{figure*}[!t]
\centering
\includegraphics[width=5.5in]{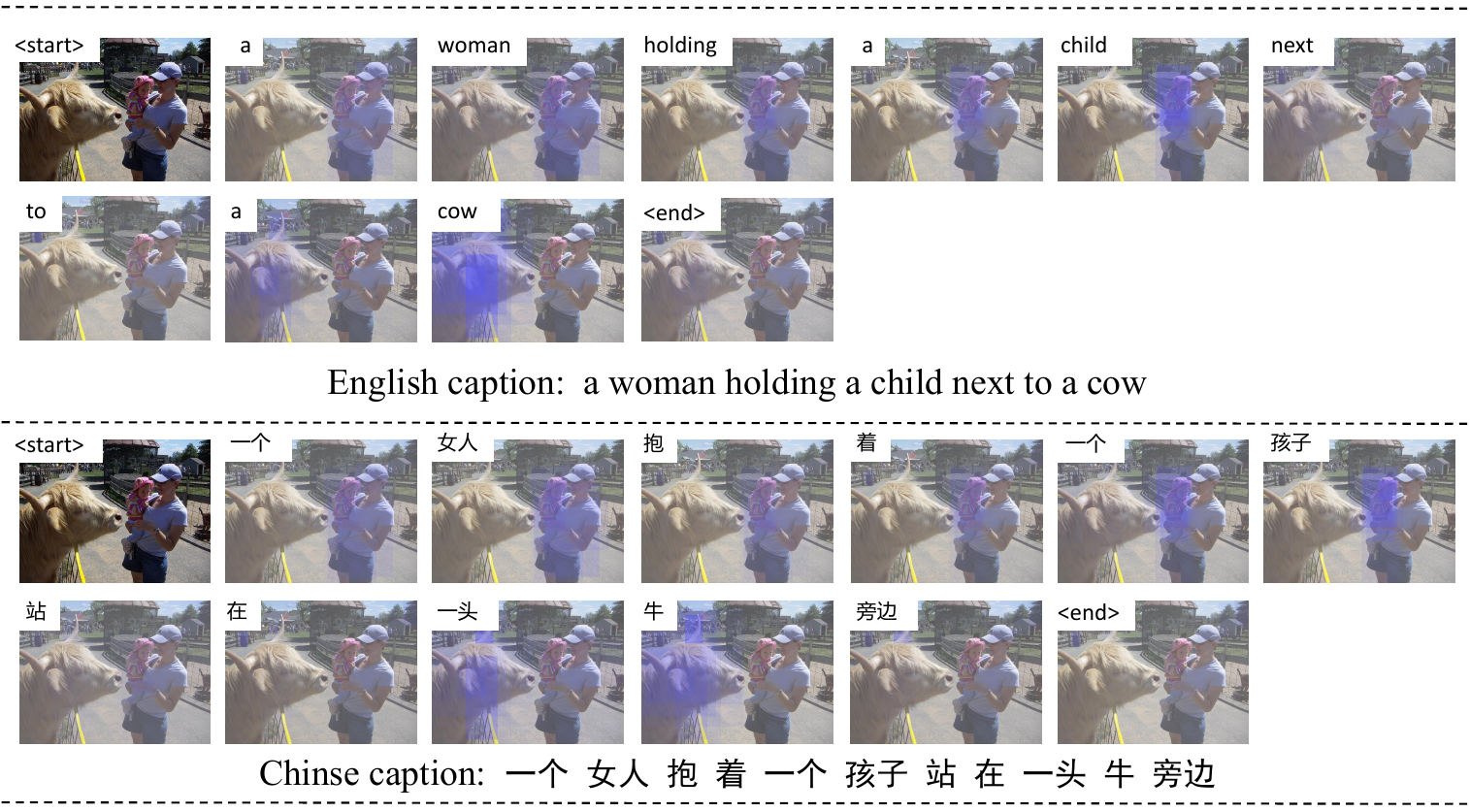}
\caption{Visualization of heterogeneous attention effect. The examples show that the attention weights of English and Chinese on the images. Each word has its own weight. The more heavily colored the areais, the more relevant it is.}
\label{Vis}
\end{figure*}
Fig.~\ref{Vis} shows the attention weights of English and Chinese words. Strong similarities in word attention can be clearly found where differences in language have almost no effect. The consistency of objects such as 'women/child/cow' are mapped into the same attention region and weight for English and Chinese captions. Additionally, articles such as 'a' are also accurately positioned. In contrast to the translation model to pursue extreme correspondence, our model allows Chinese and English to describe themselves loosely with differences where there is no 'standing' in English caption while there is in Chinese.

\begin{figure*}[!t]
\centering
\includegraphics[width=6.5in]{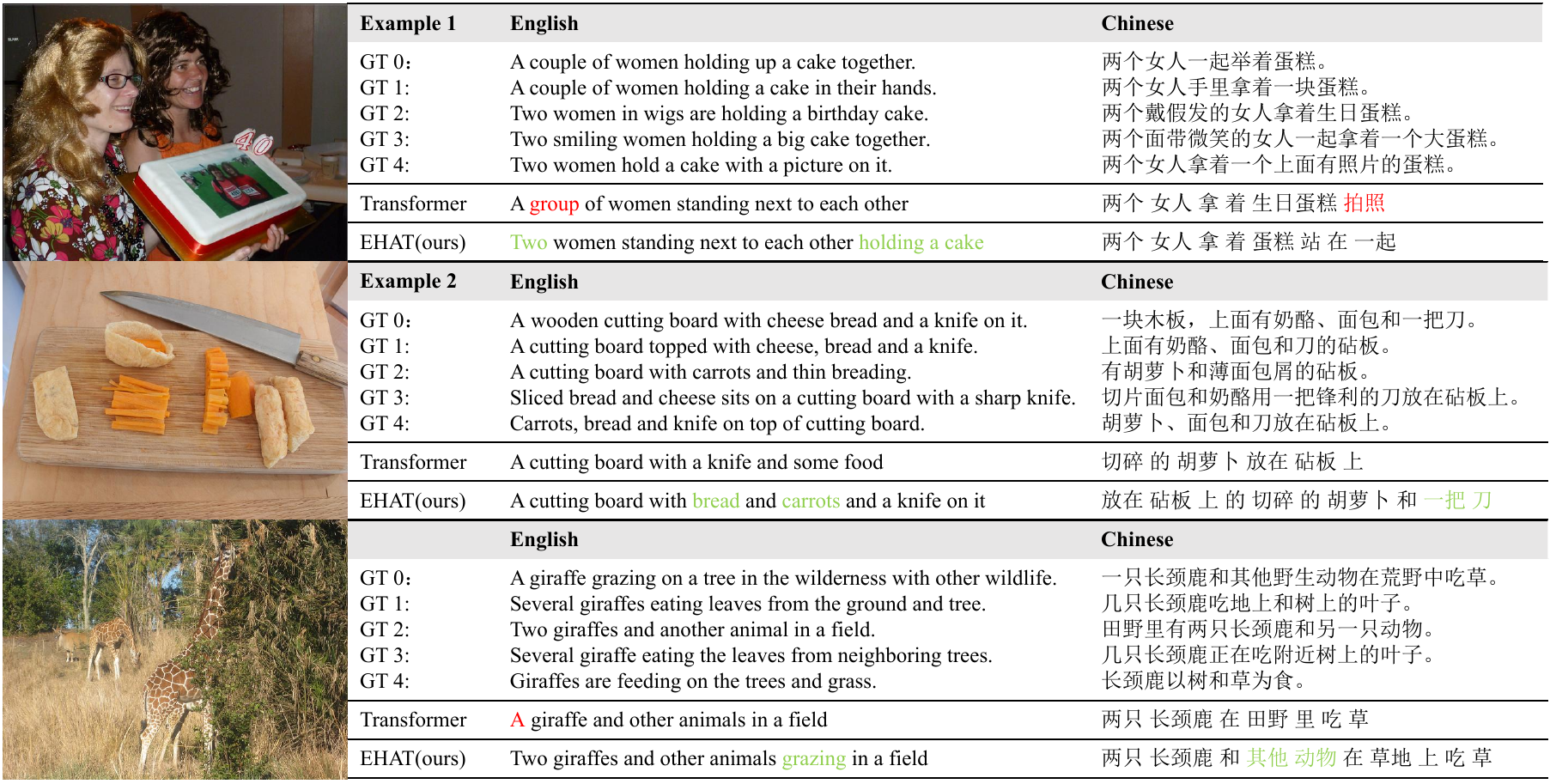}
\caption{Examples of the generated captions from transformer outputs and EHAT. GT denotes the ground-truth. We marked the words closer to the ground-truth in green font, and the words inconsistent with the ground-truth in red font.}
\label{Exam}
\end{figure*}

We also present three examples of generated image captions in Fig.~\ref{Exam}. We can intuitively understand that the effect of the EHAT not only becomes uniform in English and Chinese captions, but also learns more information from images by building heterogeneous attention. As shown in the Example~1, the captions of transformer in English and Chinese are not synchronized, which embodies in error count of women and the loss of cake in English part. As shown in the Example~2 and Example~3, our EHAT notices more key objects in the image and maintains the coordination of bilingual captions.

\subsection{Variant Results and Related Discussion}
\begin{figure*}[!t]
\centering
\includegraphics[width=5in]{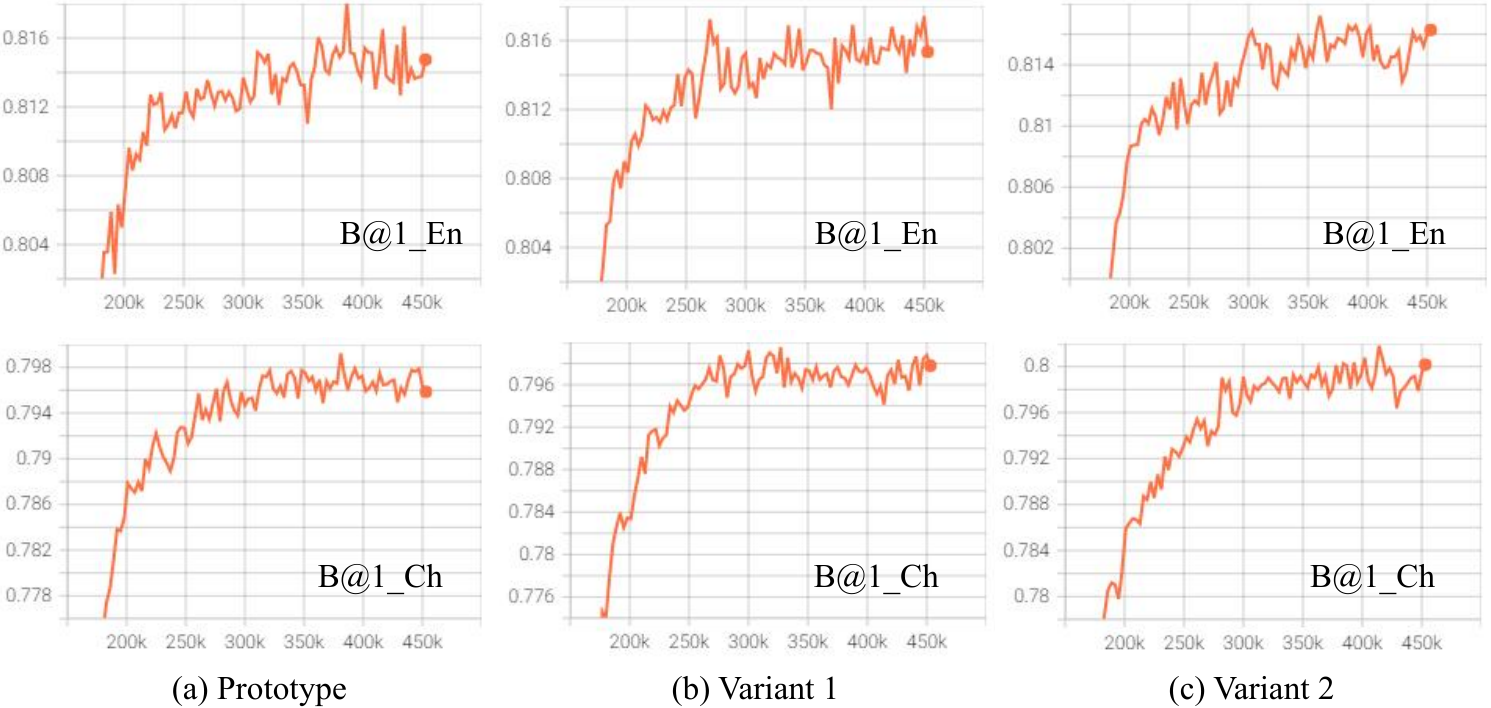}
\caption{Second training stage with 25 epochs of the B@1 metric in English and Chinese for HARN and its variants.}
\label{B1}
\end{figure*}

\begin{figure*}[!t]
\centering
\includegraphics[width=5in]{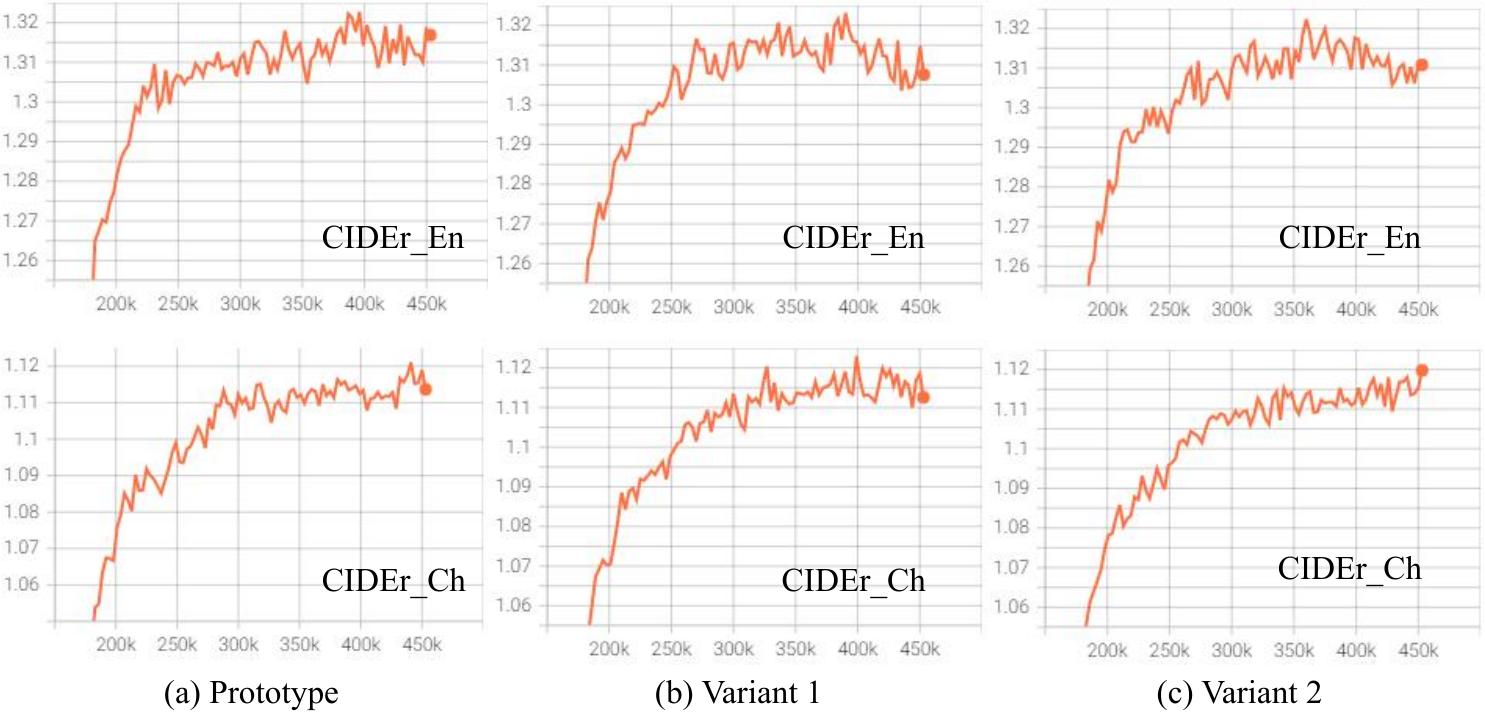}
\caption{Second training stage with 25 epochs of CIDEr metric in English and Chinese for HARN and its variants.}
\label{CIDEr}
\end{figure*}
In this section, we compare the performance and training course between the prototype and the variants to explore the interactions between languages. The experimental results are obtained from the second training stage with a 1e-5 learning rate for 25 epochs, where we increase enough training epochs to make it converge. The variants are only tweaks of the operational structure of the forward training and the number of parameters are exactly the same, which can be found in Section~\ref{sec3}.D. Thus, the results are relatively close and we display the training charts of B@1 in Fig.~\ref{B1} and CIDEr in Fig.~\ref{CIDEr}.

Variant~1 basically has a similar validation effect in training to the prototype. The English captioning B@1 exceeds 81.8 and the Chinese exceeds 79.9 at nearly 270k steps, which are faster than prototype above 100k steps. However, after 400k in CIDEr, there is a declining trend, especially for English captions. Variant~1 restricts English and Chinese to focus on the same attention maps. It can be argued that Chinese and English are close to the areas of concern for the captions. However, ignoring their differences may lead to overfitting.

For Variant~2, the Chinese B@1 improved by $80.2\%$ with bilingual interaction. In all the experiments, there is a certain gap in Chinese, which is lower than that in English. Variant~2 shows that the English corpus is beneficial for improving the Chinese corpus. However, the Chinese corpus lacks help for English intuitively in CIDEr after 360k steps.

\begin{table*}[!t]
\caption{Comparison of HARN with its variants in English and Chinese captions.}
\label{tab:var}
\centering
\begin{tabular}{c|ccccc|ccccc}
\hline
\multicolumn{1}{c|}{\multirow{2}{*}{HARN}} & \multicolumn{5}{c|}{English} &\multicolumn{5}{c}{Chinese} \\ 
\cline{2-11}
 & B@1 & B@4  & M  &R &C & B@1 & B@4  & M  &R &C\\
\hline
Prototype & \textbf{81.9} & 40.1 & \textbf{29.6} & 59.4 & \textbf{133.5} & 79.8 & \textbf{33.4} & 31.6 & 55.2 & 111.9\\
Variant 1 & 81.8 & \textbf{40.4} & \textbf{29.6} & \textbf{59.7} & 133.1 & 79.9 & 33.3 & \textbf{31.8} & \textbf{55.4} & \textbf{112.1}\\
Variant 2 & 81.7 & 39.9 & 29.5 & 59.5 & 132.7 & \textbf{80.2} & 33.1 & \textbf{31.8} & 55.3 & 111.6\\
\hline
\end{tabular}
\end{table*}

This is just an exploration of language interaction in heterogeneous learning and indicates that the heterogeneous structure is not unique. The results of the evaluation of the test set for the three structures at this training stage are listed in Table~\ref{tab:var}. We emphasize that the core of this paper is the ability of heterogeneous attention modeling for cross-lingual image captioning. To avoid interference and randomness, we chose the prototype as our main experimental subject, whether or not the variants achieves better performance.

\section{Conclusion}
\label{sec5}
We propose a Embedded Heterogeneous Attention Transform (EHAT) for cross-lingual image captioning. As we known, it is the first attempt to use heterogeneous attention modeling in such cross-lingual and cross-modal tasks. The EHAT builds reasoning paths and is focused on local matching to bridge visual features with word embedding. In addition, the heterogeneous attention embedded into the transformer can be trained on large batch computations to generate more precise and fluent captions. We tested the simultaneous generation of English and Chinese captions, which are the two most commonly used languages, but their language families are very different. The experimental results demonstrate the effectiveness of our approach for cross-lingual image captioning task and heterogeneous attention can be used to effectively model language interactions via variant experiments involving multimodal alignment. 

For potential future research, we note that current methods for cross-lingual image captioning are still based on supporting bilingual output. The possible reasons are mainly the lack of multilingual datasets and differences between different language families. To be objective, heterogeneous attention can effectively resolve differences and align representations. However, the computational complexity increases as a power function due to pairwise calculations of similarity. Thus, expanding multilingual datasets and multilingual outputs are still challenging directions for the future.


\section*{Acknowledgments}
This work was supported by the NSFC NO. 62172138 and 61932009.

\bibliographystyle{IEEEtran}
\bibliography{EHAT}

\begin{thebibliography}{10}
\providecommand{\url}[1]{#1}
\csname url@samestyle\endcsname
\providecommand{\newblock}{\relax}
\providecommand{\bibinfo}[2]{#2}
\providecommand{\BIBentrySTDinterwordspacing}{\spaceskip=0pt\relax}
\providecommand{\BIBentryALTinterwordstretchfactor}{4}
\providecommand{\BIBentryALTinterwordspacing}{\spaceskip=\fontdimen2\font plus
\BIBentryALTinterwordstretchfactor\fontdimen3\font minus
  \fontdimen4\font\relax}
\providecommand{\BIBforeignlanguage}[2]{{%
\expandafter\ifx\csname l@#1\endcsname\relax
\typeout{** WARNING: IEEEtran.bst: No hyphenation pattern has been}%
\typeout{** loaded for the language `#1'. Using the pattern for}%
\typeout{** the default language instead.}%
\else
\language=\csname l@#1\endcsname
\fi
#2}}
\providecommand{\BIBdecl}{\relax}
\BIBdecl

\bibitem{chua2009nus}
T.-S. Chua, J.~Tang, R.~Hong, H.~Li, Z.~Luo, and Y.~Zheng, ``Nus-wide: a
  real-world web image database from national university of singapore,'' in
  \emph{Proceedings of the ACM international conference on image and video
  retrieval}, 2009, pp. 1--9.

\bibitem{xie2013picture}
L.~Xie and X.~He, ``Picture tags and world knowledge: Learning tag relations
  from visual semantic sources,'' in \emph{Proceedings of the 21st ACM
  international conference on Multimedia}, 2013, pp. 967--976.

\bibitem{cui2017general}
P.~Cui, S.~Liu, and W.~Zhu, ``General knowledge embedded image representation
  learning,'' \emph{IEEE Transactions on Multimedia}, vol.~20, no.~1, pp.
  198--207, 2017.

\bibitem{chen2015microsoft}
X.~Chen, H.~Fang, T.-Y. Lin, R.~Vedantam, S.~Gupta, P.~Doll{\'a}r, and C.~L.
  Zitnick, ``Microsoft coco captions: Data collection and evaluation server,''
  \emph{arXiv preprint arXiv:1504.00325}, 2015.

\bibitem{rennie2017self}
S.~J. Rennie, E.~Marcheret, Y.~Mroueh, J.~Ross, and V.~Goel, ``Self-critical
  sequence training for image captioning,'' in \emph{Proceedings of the IEEE
  conference on computer vision and pattern recognition}, 2017, pp. 7008--7024.

\bibitem{pan2020x}
Y.~Pan, T.~Yao, Y.~Li, and T.~Mei, ``X-linear attention networks for image
  captioning,'' in \emph{Proceedings of the IEEE/CVF conference on computer
  vision and pattern recognition}, 2020, pp. 10\,971--10\,980.

\bibitem{yang2020auto}
X.~Yang, H.~Zhang, and J.~Cai, ``Auto-encoding and distilling scene graphs for
  image captioning,'' \emph{IEEE Transactions on Pattern Analysis and Machine
  Intelligence}, vol.~44, no.~5, pp. 2313--2327, 2020.

\bibitem{nguyen2022grit}
V.-Q. Nguyen, M.~Suganuma, and T.~Okatani, ``Grit: Faster and better image
  captioning transformer using dual visual features,'' in \emph{European
  Conference on Computer Vision}, 2022, pp. 167--184.

\bibitem{li2019coco}
X.~Li, C.~Xu, X.~Wang, W.~Lan, Z.~Jia, G.~Yang, and J.~Xu, ``Coco-cn for
  cross-lingual image tagging, captioning, and retrieval,'' \emph{IEEE
  Transactions on Multimedia}, vol.~21, no.~9, pp. 2347--2360, 2019.

\bibitem{miyazaki2016cross}
T.~Miyazaki and N.~Shimizu, ``Cross-lingual image caption generation,'' in
  \emph{Proceedings of the 54th Annual Meeting of the Association for
  Computational Linguistics (Volume 1: Long Papers)}, 2016, pp. 1780--1790.

\bibitem{elliott2015multilingual}
D.~Elliott, S.~Frank, and E.~Hasler, ``Multilingual image description with
  neural sequence models,'' \emph{arXiv preprint arXiv:1510.04709}, 2015.

\bibitem{anderson2018bottom}
P.~Anderson, X.~He, C.~Buehler, D.~Teney, M.~Johnson, S.~Gould, and L.~Zhang,
  ``Bottom-up and top-down attention for image captioning and visual question
  answering,'' in \emph{Proceedings of the IEEE conference on computer vision
  and pattern recognition}, 2018, pp. 6077--6086.

\bibitem{chen2021towards}
A.~Chen, X.~Huang, H.~Lin, and X.~Li, ``Towards annotation-free evaluation of
  cross-lingual image captioning,'' in \emph{Proceedings of the 2nd ACM
  International Conference on Multimedia in Asia}, 2021, pp. 1--7.

\bibitem{wang2020cross}
B.~Wang, C.~Wang, Q.~Zhang, Y.~Su, Y.~Wang, and Y.~Xu, ``Cross-lingual image
  caption generation based on visual attention model,'' \emph{IEEE Access},
  vol.~8, pp. 104\,543--104\,554, 2020.

\bibitem{huang2019attention}
L.~Huang, W.~Wang, J.~Chen, and X.-Y. Wei, ``Attention on attention for image
  captioning,'' in \emph{Proceedings of the IEEE/CVF international conference
  on computer vision}, 2019, pp. 4634--4643.

\bibitem{jaffe2017generating}
A.~Jaffe, ``Generating image descriptions using multilingual data,'' in
  \emph{Proceedings of the second conference on machine translation}, 2017, pp.
  458--464.

\bibitem{gu2018unpaired}
J.~Gu, S.~Joty, J.~Cai, and G.~Wang, ``Unpaired image captioning by language
  pivoting,'' in \emph{Proceedings of the European Conference on Computer
  Vision (ECCV)}, 2018, pp. 503--519.

\bibitem{vinyals2015show}
O.~Vinyals, A.~Toshev, S.~Bengio, and D.~Erhan, ``Show and tell: A neural image
  caption generator,'' in \emph{Proceedings of the IEEE conference on computer
  vision and pattern recognition}, 2015, pp. 3156--3164.

\bibitem{xu2015show}
K.~Xu, J.~Ba, R.~Kiros, K.~Cho, A.~Courville, R.~Salakhudinov, R.~Zemel, and
  Y.~Bengio, ``Show, attend and tell: Neural image caption generation with
  visual attention,'' in \emph{International conference on machine
  learning}.\hskip 1em plus 0.5em minus 0.4em\relax PMLR, 2015, pp. 2048--2057.

\bibitem{jiang2018recurrent}
W.~Jiang, L.~Ma, Y.-G. Jiang, W.~Liu, and T.~Zhang, ``Recurrent fusion network
  for image captioning,'' in \emph{Proceedings of the European conference on
  computer vision (ECCV)}, 2018, pp. 499--515.

\bibitem{yao2018exploring}
T.~Yao, Y.~Pan, Y.~Li, and T.~Mei, ``Exploring visual relationship for image
  captioning,'' in \emph{Proceedings of the European conference on computer
  vision (ECCV)}, 2018, pp. 684--699.

\bibitem{vaswani2017attention}
A.~Vaswani, N.~Shazeer, N.~Parmar, J.~Uszkoreit, L.~Jones, A.~N. Gomez,
  {\L}.~Kaiser, and I.~Polosukhin, ``Attention is all you need,''
  \emph{Advances in neural information processing systems}, vol.~30, 2017.

\bibitem{lu2019vilbert}
J.~Lu, D.~Batra, D.~Parikh, and S.~Lee, ``Vilbert: Pretraining task-agnostic
  visiolinguistic representations for vision-and-language tasks,''
  \emph{Advances in neural information processing systems}, vol.~32, 2019.

\bibitem{li2020oscar}
X.~Li, X.~Yin, C.~Li, P.~Zhang, X.~Hu, L.~Zhang, L.~Wang, H.~Hu, L.~Dong,
  F.~Wei \emph{et~al.}, ``Oscar: Object-semantics aligned pre-training for
  vision-language tasks,'' in \emph{Computer Vision--ECCV 2020: 16th European
  Conference, Glasgow, UK, August 23--28, 2020, Proceedings, Part XXX
  16}.\hskip 1em plus 0.5em minus 0.4em\relax Springer, 2020, pp. 121--137.

\bibitem{radford2021learning}
A.~Radford, J.~W. Kim, C.~Hallacy, A.~Ramesh, G.~Goh, S.~Agarwal, G.~Sastry,
  A.~Askell, P.~Mishkin, J.~Clark \emph{et~al.}, ``Learning transferable visual
  models from natural language supervision,'' in \emph{International conference
  on machine learning}.\hskip 1em plus 0.5em minus 0.4em\relax PMLR, 2021, pp.
  8748--8763.

\bibitem{alayrac2022flamingo}
J.-B. Alayrac, J.~Donahue, P.~Luc, A.~Miech, I.~Barr, Y.~Hasson, K.~Lenc,
  A.~Mensch, K.~Millican, M.~Reynolds \emph{et~al.}, ``Flamingo: a visual
  language model for few-shot learning,'' \emph{Advances in Neural Information
  Processing Systems}, vol.~35, pp. 23\,716--23\,736, 2022.

\bibitem{yu2021heterogeneous}
T.~Yu, Y.~Yang, Y.~Li, L.~Liu, H.~Fei, and P.~Li, ``Heterogeneous attention
  network for effective and efficient cross-modal retrieval,'' in
  \emph{Proceedings of the 44th International ACM SIGIR Conference on Research
  and Development in Information Retrieval}, 2021, pp. 1146--1156.

\bibitem{zhang2021vinvl}
P.~Zhang, X.~Li, X.~Hu, J.~Yang, L.~Zhang, L.~Wang, Y.~Choi, and J.~Gao,
  ``Vinvl: Revisiting visual representations in vision-language models,'' in
  \emph{Proceedings of the IEEE/CVF Conference on Computer Vision and Pattern
  Recognition}, 2021, pp. 5579--5588.

\bibitem{yang2018multitask}
M.~Yang, W.~Zhao, W.~Xu, Y.~Feng, Z.~Zhao, X.~Chen, and K.~Lei, ``Multitask
  learning for cross-domain image captioning,'' \emph{IEEE Transactions on
  Multimedia}, vol.~21, no.~4, pp. 1047--1061, 2018.

\bibitem{tan2019comic}
J.~H. Tan, C.~S. Chan, and J.~H. Chuah, ``Comic: Toward a compact image
  captioning model with attention,'' \emph{IEEE Transactions on Multimedia},
  vol.~21, no.~10, pp. 2686--2696, 2019.

\bibitem{xiao2019deep}
X.~Xiao, L.~Wang, K.~Ding, S.~Xiang, and C.~Pan, ``Deep hierarchical
  encoder--decoder network for image captioning,'' \emph{IEEE Transactions on
  Multimedia}, vol.~21, no.~11, pp. 2942--2956, 2019.

\bibitem{zhou2020more}
Y.~Zhou, M.~Wang, D.~Liu, Z.~Hu, and H.~Zhang, ``More grounded image captioning
  by distilling image-text matching model,'' in \emph{Proceedings of the
  IEEE/CVF conference on computer vision and pattern recognition}, 2020, pp.
  4777--4786.

\bibitem{zhang2020integrating}
J.~Zhang, K.~Mei, Y.~Zheng, and J.~Fan, ``Integrating part of speech guidance
  for image captioning,'' \emph{IEEE Transactions on Multimedia}, vol.~23, pp.
  92--104, 2020.

\bibitem{yang2020captionnet}
L.~Yang, H.~Wang, P.~Tang, and Q.~Li, ``Captionnet: A tailor-made recurrent
  neural network for generating image descriptions,'' \emph{IEEE Transactions
  on Multimedia}, vol.~23, pp. 835--845, 2020.

\bibitem{guo2020normalized}
L.~Guo, J.~Liu, X.~Zhu, P.~Yao, S.~Lu, and H.~Lu, ``Normalized and
  geometry-aware self-attention network for image captioning,'' in
  \emph{Proceedings of the IEEE/CVF conference on computer vision and pattern
  recognition}, 2020, pp. 10\,327--10\,336.

\bibitem{zhou2021semi}
Y.~Zhou, Y.~Zhang, Z.~Hu, and M.~Wang, ``Semi-autoregressive transformer for
  image captioning,'' in \emph{Proceedings of the IEEE/CVF International
  Conference on Computer Vision}, 2021, pp. 3139--3143.

\bibitem{yang2021deconfounded}
X.~Yang, H.~Zhang, and J.~Cai, ``Deconfounded image captioning: A causal
  retrospect,'' \emph{IEEE Transactions on Pattern Analysis and Machine
  Intelligence}, 2021.

\bibitem{huang2021image}
Q.~Huang, Y.~Liang, J.~Wei, Y.~Cai, H.~Liang, H.-f. Leung, and Q.~Li, ``Image
  difference captioning with instance-level fine-grained feature
  representation,'' \emph{IEEE Transactions on Multimedia}, vol.~24, pp.
  2004--2017, 2021.

\bibitem{zhang2021exploring}
Z.~Zhang, Q.~Wu, Y.~Wang, and F.~Chen, ``Exploring pairwise relationships
  adaptively from linguistic context in image captioning,'' \emph{IEEE
  Transactions on Multimedia}, 2021.

\bibitem{yu2021dual}
L.~Yu, J.~Zhang, and Q.~Wu, ``Dual attention on pyramid feature maps for image
  captioning,'' \emph{IEEE Transactions on Multimedia}, vol.~24, pp.
  1775--1786, 2021.

\bibitem{wang2022end}
Y.~Wang, J.~Xu, and Y.~Sun, ``End-to-end transformer based model for image
  captioning,'' in \emph{Proceedings of the AAAI Conference on Artificial
  Intelligence}, vol.~36, no.~3, 2022, pp. 2585--2594.

\bibitem{zhou2022compact}
Y.~Zhou, Z.~Hu, D.~Liu, H.~Ben, and M.~Wang, ``Compact bidirectional
  transformer for image captioning,'' \emph{arXiv preprint arXiv:2201.01984},
  2022.

\bibitem{wang2022text}
D.~Wang, Z.~Hu, Y.~Zhou, R.~Hong, and M.~Wang, ``A text-guided generation and
  refinement model for image captioning,'' \emph{IEEE Transactions on
  Multimedia}, 2022.

\bibitem{liu2021swin}
Z.~Liu, Y.~Lin, Y.~Cao, H.~Hu, Y.~Wei, Z.~Zhang, S.~Lin, and B.~Guo, ``Swin
  transformer: Hierarchical vision transformer using shifted windows,'' in
  \emph{Proceedings of the IEEE/CVF international conference on computer
  vision}, 2021, pp. 10\,012--10\,022.

\bibitem{lan2017fluency}
W.~Lan, X.~Li, and J.~Dong, ``Fluency-guided cross-lingual image captioning,''
  in \emph{Proceedings of the 25th ACM international conference on Multimedia},
  2017, pp. 1549--1557.

\bibitem{gao2022unison}
J.~Gao, Y.~Zhou, L.~Philip, S.~Joty, and J.~Gu, ``Unison: Unpaired
  cross-lingual image captioning,'' in \emph{Proceedings of the AAAI Conference
  on Artificial Intelligence}, vol.~36, no.~10, 2022, pp. 10\,654--10\,662.

\bibitem{dong2020cross}
W.~Dong, M.~Otani, N.~Garcia, Y.~Nakashima, and C.~Chu, ``Cross-lingual visual
  grounding,'' \emph{IEEE Access}, vol.~9, pp. 349--358, 2020.

\bibitem{chen2022cross}
Z.~Chen, F.~Yin, Q.~Yang, and C.-L. Liu, ``Cross-lingual text image recognition
  via multi-hierarchy cross-modal mimic,'' \emph{IEEE Transactions on
  Multimedia}, 2022.

\bibitem{aggarwal2021towards}
P.~Aggarwal, R.~Tambi, and A.~Kale, ``Towards zero-shot cross-lingual image
  retrieval and tagging,'' \emph{arXiv preprint arXiv:2109.07622}, 2021.

\bibitem{plummer2015flickr30k}
B.~A. Plummer, L.~Wang, C.~M. Cervantes, J.~C. Caicedo, J.~Hockenmaier, and
  S.~Lazebnik, ``Flickr30k entities: Collecting region-to-phrase
  correspondences for richer image-to-sentence models,'' in \emph{Proceedings
  of the IEEE international conference on computer vision}, 2015, pp.
  2641--2649.

\bibitem{li2016adding}
X.~Li, W.~Lan, J.~Dong, and H.~Liu, ``Adding chinese captions to images,'' in
  \emph{Proceedings of the 2016 ACM on international conference on multimedia
  retrieval}, 2016, pp. 271--275.

\bibitem{tsutsui2017using}
S.~Tsutsui and D.~Crandall, ``Using artificial tokens to control languages for
  multilingual image caption generation,'' \emph{arXiv preprint
  arXiv:1706.06275}, 2017.

\bibitem{jia2020icap}
Z.~Jia and X.~Li, ``icap: Interactive image captioning with predictive text,''
  in \emph{Proceedings of the 2020 International Conference on Multimedia
  Retrieval}, 2020, pp. 428--435.

\bibitem{stefanini2021show}
M.~Stefanini, M.~Cornia, L.~Baraldi, S.~Cascianelli, G.~Fiameni, and
  R.~Cucchiara, ``From show to tell: A survey on image captioning,''
  \emph{arXiv preprint arXiv:2107.06912}, 2021.

\bibitem{he2021synchronous}
H.~He, Q.~Wang, Z.~Yu, Y.~Zhao, J.~Zhang, and C.~Zong, ``Synchronous
  interactive decoding for multilingual neural machine translation,'' in
  \emph{Proceedings of the AAAI Conference on Artificial Intelligence},
  vol.~35, no.~14, 2021, pp. 12\,981--12\,988.

\bibitem{wang2019synchronously}
Y.~Wang, J.~Zhang, L.~Zhou, Y.~Liu, and C.~Zong, ``Synchronously generating two
  languages with interactive decoding,'' in \emph{Proceedings of the 2019
  Conference on Empirical Methods in Natural Language Processing and the 9th
  International Joint Conference on Natural Language Processing
  (EMNLP-IJCNLP)}, 2019, pp. 3350--3355.

\bibitem{zhou2021uc2}
M.~Zhou, L.~Zhou, S.~Wang, Y.~Cheng, L.~Li, Z.~Yu, and J.~Liu, ``Uc2: Universal
  cross-lingual cross-modal vision-and-language pre-training,'' in
  \emph{Proceedings of the IEEE/CVF Conference on Computer Vision and Pattern
  Recognition}, 2021, pp. 4155--4165.

\bibitem{sun2011pathsim}
Y.~Sun, J.~Han, X.~Yan, P.~S. Yu, and T.~Wu, ``Pathsim: Meta path-based top-k
  similarity search in heterogeneous information networks,'' \emph{Proceedings
  of the VLDB Endowment}, vol.~4, no.~11, pp. 992--1003, 2011.

\bibitem{sun2012mining}
Y.~Sun and J.~Han, ``Mining heterogeneous information networks: principles and
  methodologies,'' \emph{Synthesis Lectures on Data Mining and Knowledge
  Discovery}, vol.~3, no.~2, pp. 1--159, 2012.

\bibitem{shi2018heterogeneous}
C.~Shi, B.~Hu, W.~X. Zhao, and S.~Y. Philip, ``Heterogeneous information
  network embedding for recommendation,'' \emph{IEEE Transactions on Knowledge
  and Data Engineering}, vol.~31, no.~2, pp. 357--370, 2018.

\bibitem{hu2018leveraging}
B.~Hu, C.~Shi, W.~X. Zhao, and P.~S. Yu, ``Leveraging meta-path based context
  for top-n recommendation with a neural co-attention model,'' in
  \emph{Proceedings of the 24th ACM SIGKDD international conference on
  knowledge discovery \& data mining}, 2018, pp. 1531--1540.

\bibitem{fu2017hin2vec}
T.-y. Fu, W.-C. Lee, and Z.~Lei, ``Hin2vec: Explore meta-paths in heterogeneous
  information networks for representation learning,'' in \emph{Proceedings of
  the 2017 ACM on Conference on Information and Knowledge Management}, 2017,
  pp. 1797--1806.

\bibitem{hamilton2017inductive}
W.~Hamilton, Z.~Ying, and J.~Leskovec, ``Inductive representation learning on
  large graphs,'' \emph{Advances in neural information processing systems},
  vol.~30, 2017.

\bibitem{zhao2021heterogeneous}
J.~Zhao, X.~Wang, C.~Shi, B.~Hu, G.~Song, and Y.~Ye, ``Heterogeneous graph
  structure learning for graph neural networks,'' in \emph{Proceedings of the
  AAAI Conference on Artificial Intelligence}, vol.~35, no.~5, 2021, pp.
  4697--4705.

\bibitem{linmei2019heterogeneous}
H.~Linmei, T.~Yang, C.~Shi, H.~Ji, and X.~Li, ``Heterogeneous graph attention
  networks for semi-supervised short text classification,'' in
  \emph{Proceedings of the 2019 Conference on Empirical Methods in Natural
  Language Processing and the 9th International Joint Conference on Natural
  Language Processing (EMNLP-IJCNLP)}, 2019, pp. 4821--4830.

\bibitem{hu2020graph}
L.~Hu, S.~Xu, C.~Li, C.~Yang, C.~Shi, N.~Duan, X.~Xie, and M.~Zhou, ``Graph
  neural news recommendation with unsupervised preference disentanglement,'' in
  \emph{Proceedings of the 58th annual meeting of the association for
  computational linguistics}, 2020, pp. 4255--4264.

\bibitem{wang2022survey}
X.~Wang, D.~Bo, C.~Shi, S.~Fan, Y.~Ye, and S.~Y. Philip, ``A survey on
  heterogeneous graph embedding: methods, techniques, applications and
  sources,'' \emph{IEEE Transactions on Big Data}, 2022.

\bibitem{hu2020heterogeneous}
Z.~Hu, Y.~Dong, K.~Wang, and Y.~Sun, ``Heterogeneous graph transformer,'' in
  \emph{Proceedings of The Web Conference 2020}, 2020, pp. 2704--2710.

\bibitem{yao2020heterogeneous}
S.~Yao, T.~Wang, and X.~Wan, ``Heterogeneous graph transformer for
  graph-to-sequence learning,'' in \emph{Proceedings of the 58th Annual Meeting
  of the Association for Computational Linguistics}, 2020, pp. 7145--7154.

\bibitem{mei2022relation}
X.~Mei, X.~Cai, L.~Yang, and N.~Wang, ``Relation-aware heterogeneous graph
  transformer based drug repurposing,'' \emph{Expert Systems with
  Applications}, vol. 190, p. 116165, 2022.

\bibitem{zhu2020mucko}
Z.~Zhu, J.~Yu, Y.~Wang, Y.~Sun, Y.~Hu, and Q.~Wu, ``Mucko: multi-layer
  cross-modal knowledge reasoning for fact-based visual question answering,''
  \emph{arXiv preprint arXiv:2006.09073}, 2020.

\bibitem{yu2020cross}
J.~Yu, Z.~Zhu, Y.~Wang, W.~Zhang, Y.~Hu, and J.~Tan, ``Cross-modal knowledge
  reasoning for knowledge-based visual question answering,'' \emph{Pattern
  Recognition}, vol. 108, p. 107563, 2020.

\bibitem{yu2019heterogeneous}
W.~Yu, J.~Zhou, W.~Yu, X.~Liang, and N.~Xiao, ``Heterogeneous graph learning
  for visual commonsense reasoning,'' \emph{Advances in Neural Information
  Processing Systems}, vol.~32, 2019.

\bibitem{song2023efficient}
Z.~Song, Z.~Hu, and R.~Hong, ``Efficient and self-adaptive rationale knowledge
  base for visual commonsense reasoning,'' \emph{Multimedia Systems}, vol.~29,
  no.~5, pp. 3017--3026, 2023.

\bibitem{fan2019heterogeneous}
C.~Fan, X.~Zhang, S.~Zhang, W.~Wang, C.~Zhang, and H.~Huang, ``Heterogeneous
  memory enhanced multimodal attention model for video question answering,'' in
  \emph{Proceedings of the IEEE/CVF conference on computer vision and pattern
  recognition}, 2019, pp. 1999--2007.

\bibitem{jiang2020reasoning}
P.~Jiang and Y.~Han, ``Reasoning with heterogeneous graph alignment for video
  question answering,'' in \emph{Proceedings of the AAAI Conference on
  Artificial Intelligence}, vol.~34, no.~07, 2020, pp. 11\,109--11\,116.

\bibitem{cai2021heterogeneous}
D.~Cai, S.~Qian, Q.~Fang, and C.~Xu, ``Heterogeneous hierarchical feature
  aggregation network for personalized micro-video recommendation,'' \emph{IEEE
  Transactions on Multimedia}, vol.~24, pp. 805--818, 2021.

\bibitem{cai2022heterogeneous}
D.~Cai, S.~Qian, Q.~Fang, J.~Hu, W.~Ding, and C.~Xu, ``Heterogeneous graph
  contrastive learning network for personalized micro-video recommendation,''
  \emph{IEEE Transactions on Multimedia}, 2022.

\bibitem{zhu2022latent}
P.~Zhu, X.~Yao, Y.~Wang, M.~Cao, B.~Hui, S.~Zhao, and Q.~Hu, ``Latent
  heterogeneous graph network for incomplete multi-view learning,'' \emph{IEEE
  Transactions on Multimedia}, 2022.

\bibitem{gu2018stack}
J.~Gu, J.~Cai, G.~Wang, and T.~Chen, ``Stack-captioning: Coarse-to-fine
  learning for image captioning,'' in \emph{Proceedings of the AAAI Conference
  on Artificial Intelligence}, vol.~32, no.~1, 2018.

\bibitem{luo2020better}
R.~Luo, ``A better variant of self-critical sequence training,'' \emph{arXiv
  preprint arXiv:2003.09971}, 2020.

\bibitem{karpathy2015deep}
A.~Karpathy and L.~Fei-Fei, ``Deep visual-semantic alignments for generating
  image descriptions,'' in \emph{Proceedings of the IEEE conference on computer
  vision and pattern recognition}, 2015, pp. 3128--3137.

\bibitem{yang2019auto}
X.~Yang, K.~Tang, H.~Zhang, and J.~Cai, ``Auto-encoding scene graphs for image
  captioning,'' in \emph{Proceedings of the IEEE/CVF Conference on Computer
  Vision and Pattern Recognition}, 2019, pp. 10\,685--10\,694.

\bibitem{herdade2019image}
S.~Herdade, A.~Kappeler, K.~Boakye, and J.~Soares, ``Image captioning:
  Transforming objects into words,'' \emph{Advances in Neural Information
  Processing Systems}, vol.~32, 2019.

\bibitem{cornia2020meshed}
M.~Cornia, M.~Stefanini, L.~Baraldi, and R.~Cucchiara, ``Meshed-memory
  transformer for image captioning,'' in \emph{Proceedings of the IEEE/CVF
  conference on computer vision and pattern recognition}, 2020, pp.
  10\,578--10\,587.

\bibitem{fan2021tcic}
Z.~Fan, Z.~Wei, S.~Wang, R.~Wang, Z.~Li, H.~Shan, and X.~Huang, ``Tcic: Theme
  concepts learning cross language and vision for image captioning,''
  \emph{arXiv preprint arXiv:2106.10936}, 2021.

\bibitem{zhang2021rstnet}
X.~Zhang, X.~Sun, Y.~Luo, J.~Ji, Y.~Zhou, Y.~Wu, F.~Huang, and R.~Ji, ``Rstnet:
  Captioning with adaptive attention on visual and non-visual words,'' in
  \emph{Proceedings of the IEEE/CVF conference on computer vision and pattern
  recognition}, 2021, pp. 15\,465--15\,474.

\bibitem{zeng2022s2}
P.~Zeng, H.~Zhang, J.~Song, and L.~Gao, ``S2 transformer for image
  captioning,'' in \emph{Proceedings of the International Joint Conferences on
  Artificial Intelligence}, vol.~5, 2022.

\end{thebibliography}

\begin{IEEEbiography}[{\includegraphics[width=1in,height=1.25in,clip,keepaspectratio]{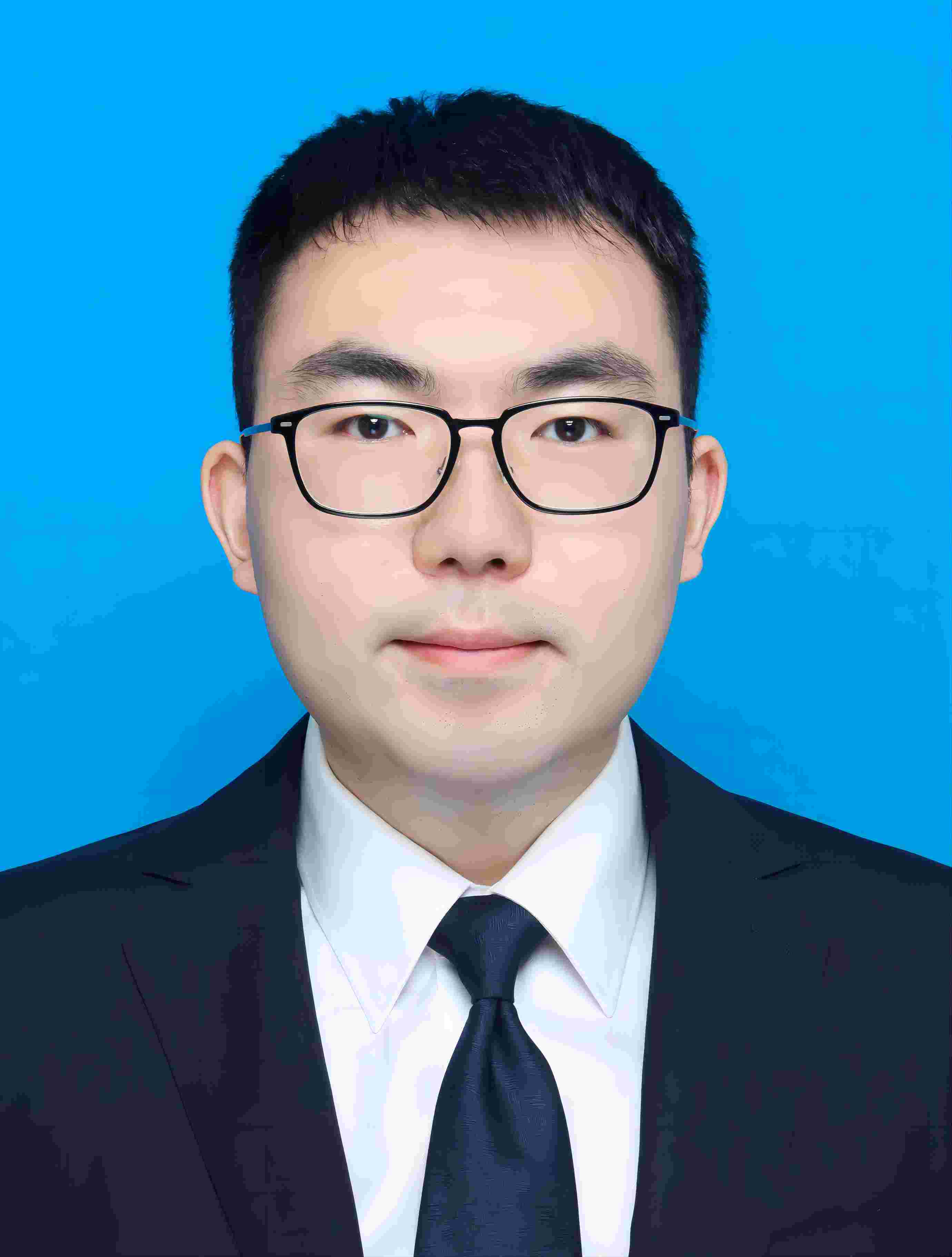}}]{Zijie Song} received the B.S. degree from North China Electric Power University, Beijing, China, in 2019 and M.E. degree from Hefei University of Technology, Hefei, China, in 2022. He is currently pursuing the Ph.D. degree in computer science and technology with the School of Computer Science and Information Engineering, Hefei University of Technology, Hefei, China. His research interests include cross-media analysis and reasoning.
\end{IEEEbiography}

\begin{IEEEbiography}[{\includegraphics[width=1in,height=1.25in,clip,keepaspectratio]{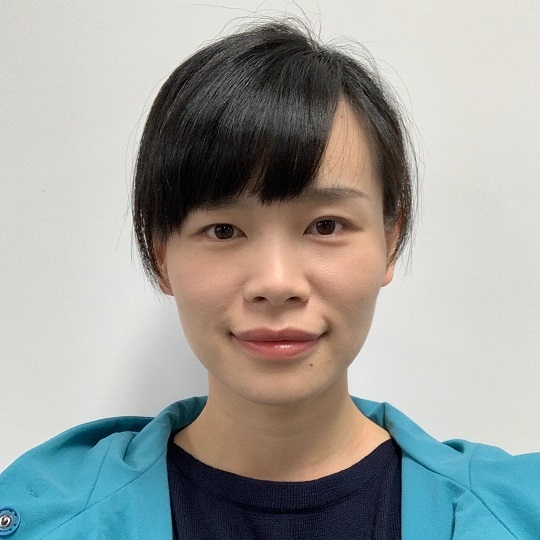}}]{Zhenzhen Hu} Dr. Zhenzhen Hu is an Associate Professor with School of Computer and Information at Hefei University of Technology (HFUT), China. She received her PhD degree from HFUT in 2014, under the supervision of Prof. Jianguo Jiang and Prof. Richang Hong. She used to be a Research Fellow in at Cloud Computing Application and Platform Group of Nanyang Technological University, directed by Prof. Yonggang Wen. Her current research interests include cross-media computing and computer vision.
\end{IEEEbiography}

\begin{IEEEbiography}[{\includegraphics[width=1in,height=1.25in,clip,keepaspectratio]{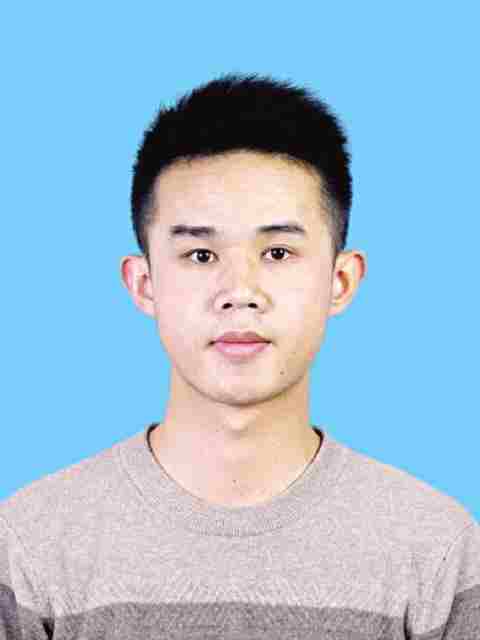}}]{Yuanen Zhou} received the Ph.D. degree from Hefei University of Technology (HUFT), Hefei, China, in 2022. He is currently a Special Associate Research Fellow of the Hefei Comprehensive National Science Center, Institute of Artificial Intelligence. His research interests include Vision\&Language, physiological signal processing, and affective computing.
\end{IEEEbiography}

\begin{IEEEbiography}[{\includegraphics[width=1in,height=1.25in,clip,keepaspectratio]{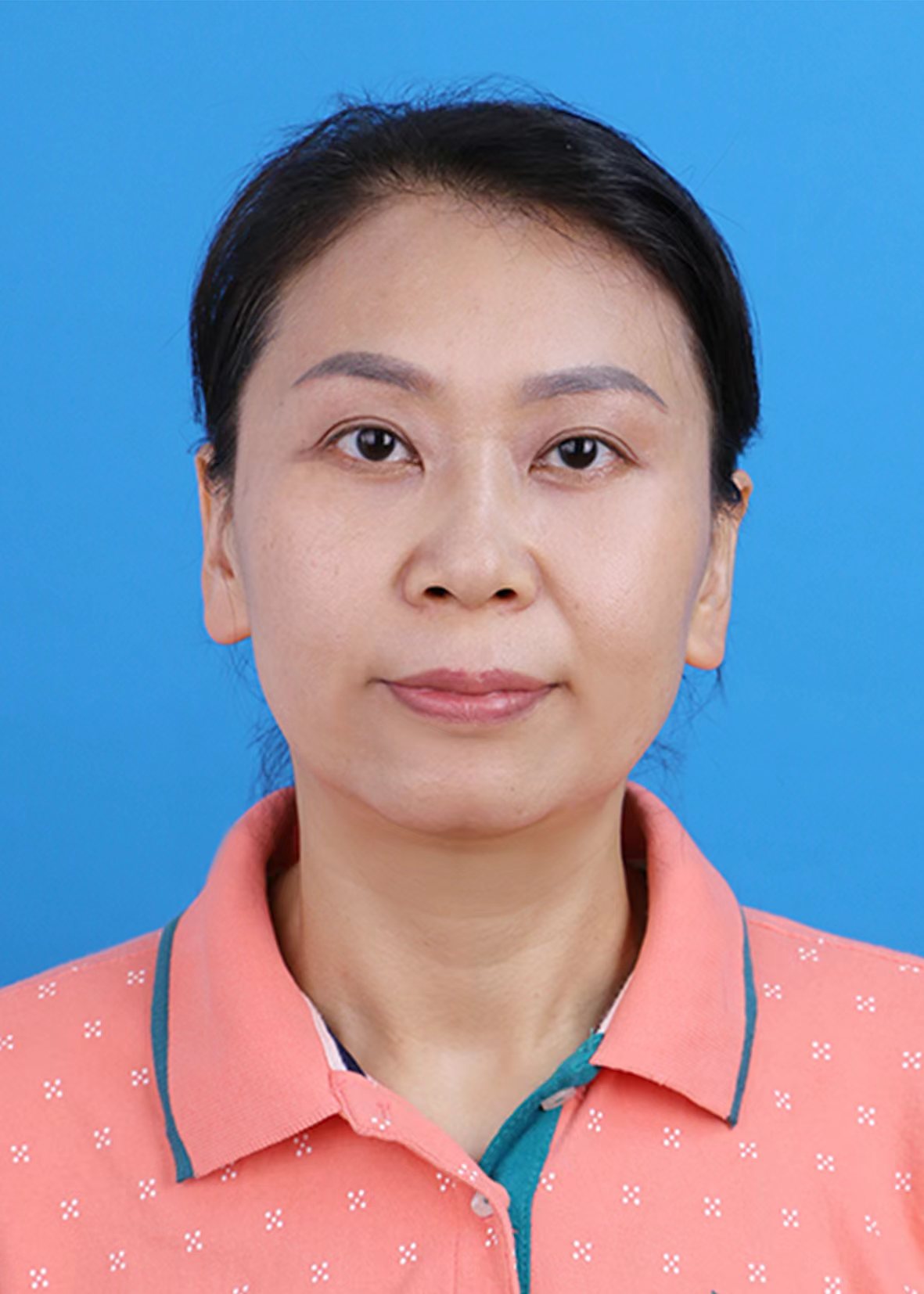}}]{Ye Zhao} received the M.S. degree in communication and information system from Harbin Engineering University, Harbin, China, in 2005 and the Ph.D. degree in signal and information processing from Hefei University of Technology, Hefei, China, in 2014.
She is an associate professor in School of Computer and Information, Hefei University of Technology. From 2016 to 2017, she was a visiting scholar in Computer Science department, University of Central Florida, USA. Her research interest includes Multimedia Analysis and Pattern Recognition.
\end{IEEEbiography}

\begin{IEEEbiography}[{\includegraphics[width=1in,height=1.25in,clip,keepaspectratio]{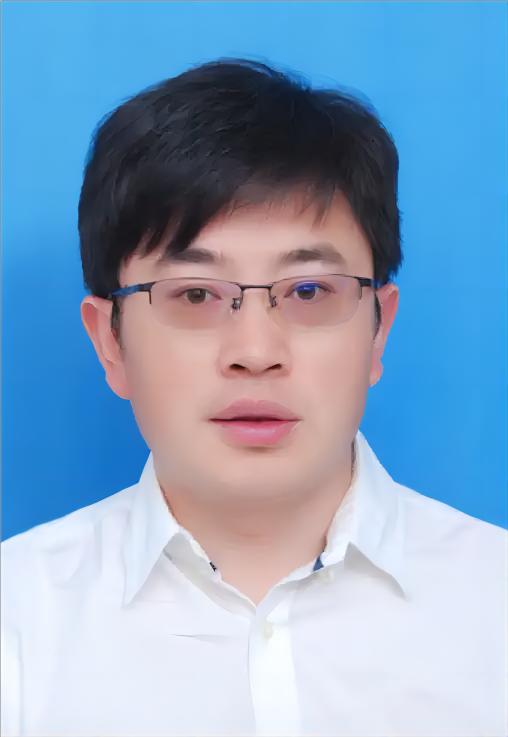}}]{Richang Hong} received the Ph.D. degree from the University of Science and Technology of China, Hefei, China, in 2008. He is currently a Professor with the Hefei University of Technology, Hefei. His research interests include multimedia content analysis and social media, in which he has coauthored more than 100 publications. He is a member of the ACM and an Executive Committee Member of the ACM SIGMM China Chapter. He was the Technical Program Chair of the MMM 2016, ICIMCS 2017, and PCM 2018. He was a recipient of the Best Paper Award at the ACM Multimedia 2010, the Best Paper Award at the ACM ICMR 2015, and the Honorable Mention of IEEE Transactions on Multimedia Best Paper Award 2015. He was an Associate Editor of IEEE Multimedia Magazine and Information Sciences and Signal Processing, Elsevier.
\end{IEEEbiography}

\begin{IEEEbiography}[{\includegraphics[width=1in,height=1.25in,clip,keepaspectratio]{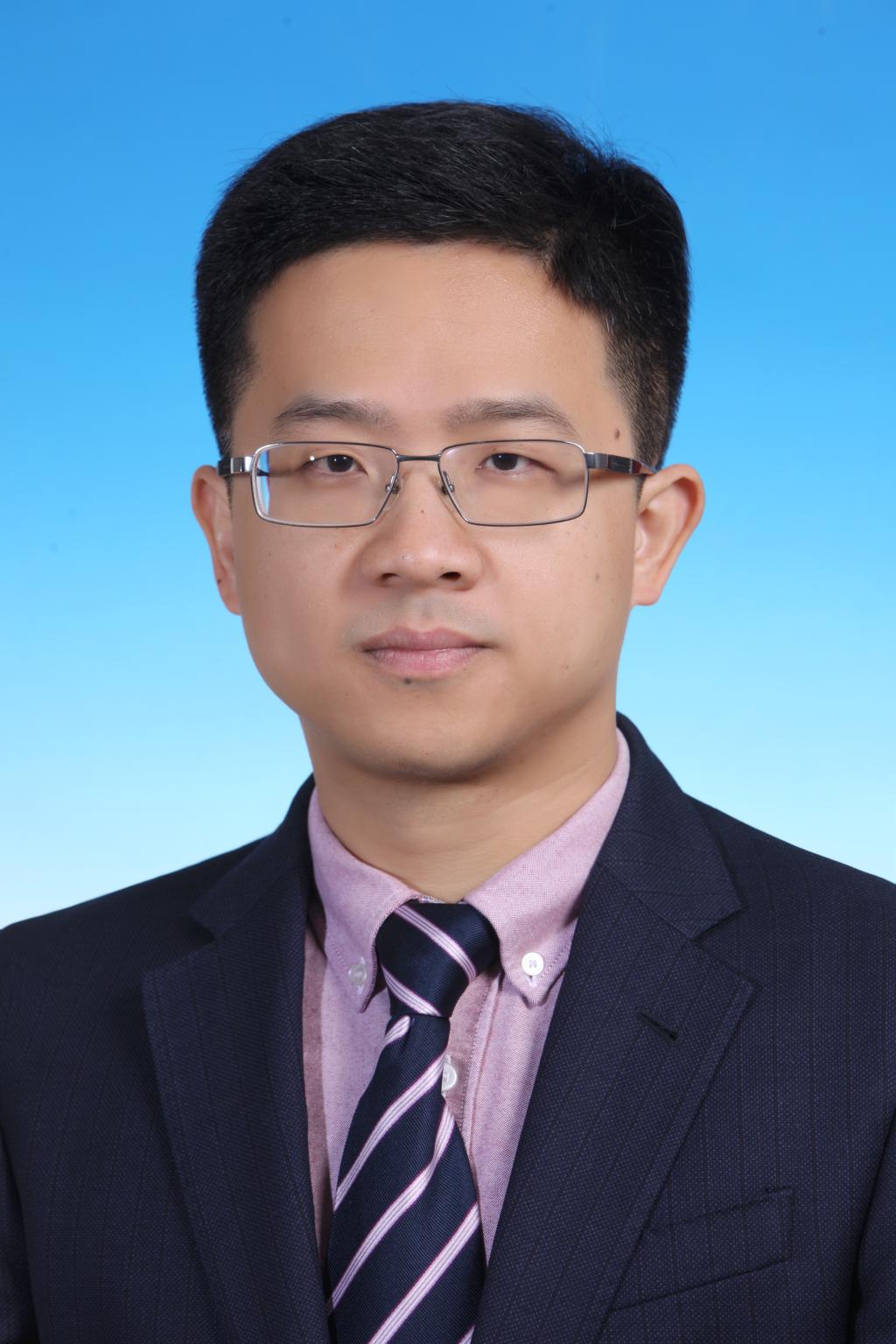}}]{Meng Wang} received the B.E. and Ph.D. degrees in signal and information processing with special class for the Gifted Young from the Department of Electronic Engineering and Information Science, University of Science and Technology of China, Hefei, China, in 2003 and 2008, respectively. He is currently a Professor with Hefei University of Technology, Hefei, China. He has authored over 200 book chapters, journals, and conference papers in his research areas. His current research interests include multimedia content analysis, computer vision, and pattern recognition. Dr. Wang was a recipient of the ACM SIGMM Rising Star Award 2014. He is an Associate Editor of IEEE Transactions on Knowledge and Data Engineering, IEEE Transactions on Circuits and Systems for Video Technology, IEEE Transactions on Multimedia, and IEEE Transactions on Neural Networks and Learning Systems.
\end{IEEEbiography}

\end{document}